\documentclass[sigconf]{acmart}

\AtBeginDocument{%
  }

\setcopyright{acmlicensed}
\copyrightyear{2026}
\acmYear{2026}
\acmDOI{XXXXXXX.XXXXXXX}

\acmConference[WWW '26]{The Web Conference 2026}{April 28--May 2,
  2026}{Sydney, Australia}
\acmISBN{978-1-4503-XXXX-X/2026/04}

\usepackage{xcolor}
\usepackage{booktabs}
\usepackage{multirow}
\usepackage{amsmath,amsfonts}
\usepackage{algorithmic}
\usepackage{algorithm}
\usepackage{array}
\usepackage[caption=false,font=normalsize,labelfont=sf,textfont=sf]{subfig}
\usepackage{textcomp}
\usepackage{verbatim}

\begin{document}

\title{MolBridge: Atom-Level Joint Graph Refinement for Robust Drug-Drug Interaction Event Prediction}

\author{Xuan Lin}
\email{jack_lin@xtu.edu.cn}
\affiliation{%
  \institution{Xiangtan University}
  \city{Xiangtan}
  \country{China}
}

\author{Aocheng Ding}
\email{angelfacedacs@stu.xmu.edu.cn}
\affiliation{%
  \institution{Xiamen University}
  \city{Xiamen}
  \country{China}
}

\author{Tengfei Ma}
\authornote{Corresponding author.}
\email{tfma@hnu.edu.cn}
\affiliation{%
  \institution{Hunan University}
  \city{Changsha}
  \country{China}
}

\author{Hua Liang}
\email{lianghua0901@hnu.edu.cn}
\affiliation{%
  \institution{Hunan University}
  \city{Changsha}
  \country{China}
}

\author{Zhe Quan}
\email{quanzhe@hnu.edu.cn}
\affiliation{%
  \institution{Hunan University}
  \city{Changsha}
  \country{China}
}

\renewcommand{\shortauthors}{Lin et al.}

\begin{abstract}
Drug combinations offer therapeutic benefits but also carry the risk of adverse drug–drug interactions (DDIs), especially under complex molecular structures. Accurate DDI event prediction requires capturing fine-grained inter-drug relationships, which are critical for modeling metabolic mechanisms such as \textit{enzyme-mediated competition}. However, existing approaches typically rely on isolated drug representations and fail to explicitly model atom-level cross-molecular interactions, limiting their effectiveness across diverse molecular complexities and DDI type distributions. To address these limitations, we propose MolBridge, a novel atom-level joint graph refinement framework for robust DDI event prediction. MolBridge constructs a joint graph that integrates atomic structures of drug pairs, enabling direct modeling of inter-drug associations. A central challenge in such joint graph settings is the potential loss of information caused by over-smoothing when modeling long-range atomic dependencies. To overcome this, we introduce a structure consistency module that iteratively refines node features while preserving the global structural context. This joint design allows MolBridge to effectively learn both local and global interaction patterns, yielding robust representations across both frequent and rare DDI types. Extensive experiments on two benchmark datasets show that MolBridge consistently 
\textcolor{black}{outperforms state-of-the-art baselines, achieving superior performance across long-tail and inductive scenarios.}
% outperforms state-of-the-art baselines, with at least 2.19\% and 2.49\% improvements in Macro-F1 and Macro-Recall, respectively. 
These results demonstrate the advantages of fine-grained graph refinement in improving the accuracy, robustness, and mechanistic interpretability of DDI event prediction. \textcolor{black}{This work contributes to Web Mining and Content Analysis by developing graph-based methods for mining and analyzing drug-drug interaction networks.} The source code is available at \url{https://anonymous.4open.science/r/MolBridge-6561}.
\end{abstract}

\begin{CCSXML}
<ccs2012>
<concept>
<concept_id>10002950.10003624.10003633.10003636</concept_id>
<concept_desc>Mathematics of computing~Graph algorithms</concept_desc>
<concept_significance>500</concept_significance>
</concept>
<concept>
<concept_id>10010147.10010257.10010258.10010259</concept_id>
<concept_desc>Computing methodologies~Neural networks</concept_desc>
<concept_significance>500</concept_significance>
</concept>
<concept>
<concept_id>10010147.10010178.10010224.10010225</concept_id>
<concept_desc>Computing methodologies~Machine learning</concept_desc>
<concept_significance>300</concept_significance>
</concept>
<concept>
<concept_id>10003120.10003121.10003129.10003131</concept_id>
<concept_desc>Human-centered computing~Bioinformatics</concept_desc>
<concept_significance>300</concept_significance>
</concept>
</ccs2012>
\end{CCSXML}

\ccsdesc[500]{Mathematics of computing~Graph algorithms}
\ccsdesc[500]{Computing methodologies~Neural networks}
\ccsdesc[300]{Computing methodologies~Machine learning}
\ccsdesc[300]{Human-centered computing~Bioinformatics}

\keywords{Drug Discovery, Drug-drug interaction prediction, Graph Neural Networks, Bioinformatics}

\maketitle

\section{Introduction}
Drug combinations have emerged as a promising strategy to enhance treatment efficacy by simultaneously targeting multiple biological pathways~\cite{han2017synergistic}. However, such combination therapy also carries an increased risk of adverse drug–drug interactions (DDIs), which can reduce efficacy or induce severe toxicity. These interactions often stem from metabolic competition, enzyme inhibition, or pharmacodynamic interference, posing significant challenges to patient safety. Therefore,
accurately predicting DDI event is critical for developing safer and more effective treatment options. More importantly, a reliable DDI event prediction framework can help mitigate adverse drug reactions, optimize clinical decisions, and ultimately improve patient outcomes~\cite{armani2024effect}

\begin{figure*}[t]
    \centering
    \includegraphics[width=1\linewidth]{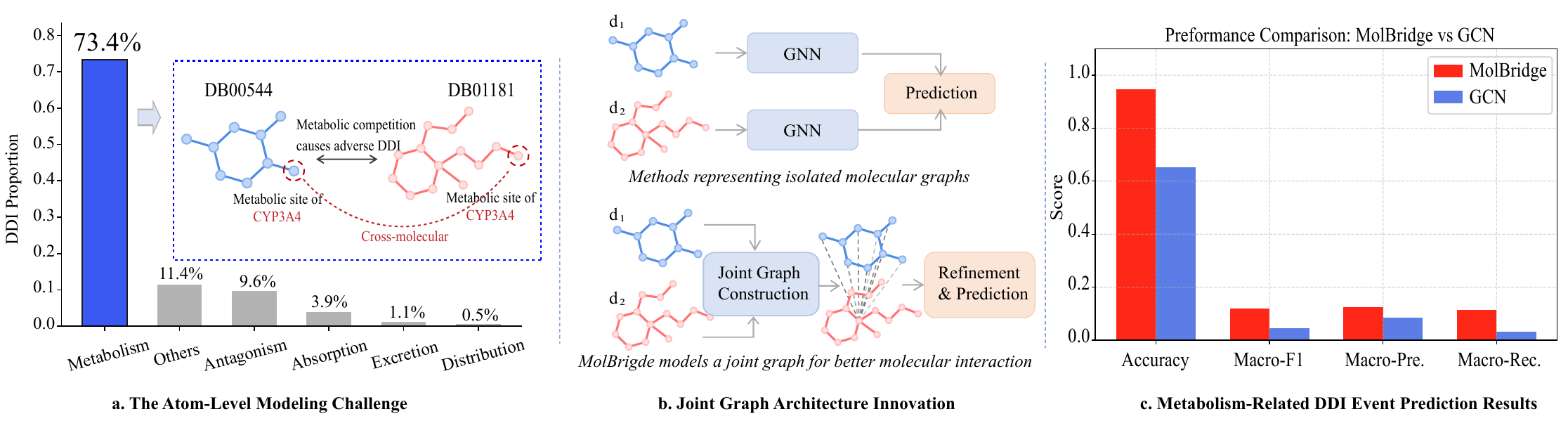}
    \caption{(a) Metabolism-related interactions account for 73.4\% of DDI types~\cite{tian2025ddinter}, yet existing GNN methods (e.g., GCN~\cite{kipf2017semisupervised}) rely on coarse drug-level representations that cannot capture metabolic DDI complexity, making fine-grained atomic-level modeling crucial. (b) Compared to GCN that treats a drug as an isolated graph, MolBridge achieves better performance by explicitly modeling cross-molecular atomic associations. (c) MolBridge comprehensively outperforms GCN on metabolism-related samples across all benchmark metrics.}
    \label{fig:intro}
\end{figure*}

Traditional wet-lab methods for DDI event prediction are often labor-intensive and costly, limiting their scalability in large-scale screening~\cite{morro2017stochastic,zitnik2018modeling}. To overcome these challenges, deep learning–based approaches have been developed to improve efficiency and predictive performance~\cite{ryu2018deep,zhong2024learning}. Among them, graph neural networks (GNNs) have shown strong potential in modeling DDIs from graph-structured data~\cite{wu2020comprehensive,li2023dsn}.
Recent efforts in graph construction for DDI event prediction mainly follow two paradigms. The first models an isolated molecular graph, where atoms are nodes and chemical bonds are edges~\cite{nyamabo2021ssi,tang2023dsil,DBLP:conf/www/WangMCW21}, capturing atom-level substructures and functional groups that often underlie DDI mechanisms. The second builds drug-centric interaction graphs, treating drugs as nodes and incorporating biomedical associations (e.g., protein targets, diseases, and side effects) as edges~\cite{wang2017knowledge,zitnik2018modeling,yu2021sumgnn,lyu2021mdnn,su2022biomedical}. These approaches effectively leverage high-level semantic context to enhance drug representation and reasoning~\cite{karim2019drug,ma2024learning}.
More recent studies attempt to combine structural and semantic information through representation fusion, aiming to improve drug modeling from both perspectives~\cite{xiong2023multi,su2024dual}. Despite these advancements, existing models still struggle to explicitly represent atom-level associations between drugs, which are essential for capturing metabolic mechanisms such as enzyme-mediated competition.
As shown in Figure~\ref{fig:intro}a, metabolism-related DDIs account for a dominant 73.4\% among all interaction types, underscoring the importance of modeling enzyme-mediated mechanisms. However, existing GNN-based methods (e.g., GCN~\cite{kipf2017semisupervised}) typically treat drugs as isolated molecular graphs (Figure~\ref{fig:intro}b), relying on coarse or drug-level representations that fail to capture the mechanistic complexity of metabolic interactions, leading to suboptimal performance on metabolism-related DDIs (Figure~\ref{fig:intro}c).

To address the aforementioned challenges, we propose MolBridge, a novel atom-based joint graph refinement framework for robust DDI event prediction. MolBridge first constructs a joint graph that encodes local atom-level relationships between paired molecular structures. Then we introduce a structure consistency module to model long-range atom-level dependencies across drug pairs on the constructed joint graph, while mitigating the over-smoothing issue common in GNN-based methods (Appendix Figure~\ref{fig:residul}). Finally, MolBridge effectively captures the full spectrum of inter-drug relationships by integrating both localized structural features and global contextual interactions. This fine-grained and hierarchical representation allows the model to better reflect underlying metabolic mechanisms (Figure~\ref{fig:intro}c), thereby achieving notable improvements in predictive accuracy and robustness across diverse DDI scenarios.
In summary, the main contributions of this work are described as follows.

\begin{itemize}
    \item We reveal that learning atom-level cross-molecular relationships provides key insight for capturing mechanistically relevant interaction patterns, further empowering the ability to generalize to diverse DDI types.
    \item We propose MolBridge, a novel framework that captures both fine-grained and long-range inter-drug associations. Central to our design is a structure consistency module, which refines joint molecular graphs by modeling long-range atom-level dependencies while mitigating the over-smoothing issues inherent in GNN-based methods.
    \item Extensive experiments on two benchmarking datasets demonstrate that MolBridge consistently outperforms state-of-the-art baselines, exhibits strong robustness in predicting rare and structurally complex DDI types, and enhances interpretability, thereby facilitating more reliable drug discovery.
\end{itemize}

\section{Related work}
\noindent\textbf{Methods Based on Local Molecular Structure.} Molecular graphs play a crucial role in encoding molecular structures, facilitating a better understanding of the chemical structures within molecules~\cite{ryu2018deep,al2022prediction}. SSI-DDI~\cite{nyamabo2021ssi} constructs a molecular graph structural model with the atoms of drug molecules as nodes and chemical bonds as edges and applies Graph Attention Network (GAT) ~\cite{velivckovic2017graph} on the graph to capture the substructure-based association of drug molecules, improving the molecular graph representation for DDI event prediction. GMPNN~\cite{nyamabo2022drug} proposes a novel gated information-passing mechanism to learn chemical sub-structures of different sizes from drug molecules for feature extraction.
Despite the effectiveness of molecular graph-based methods, it is still challenging to model atom-level cross-molecular interactions for drug representations that rely on isolated molecular graphs.

\noindent\textbf{Methods Based on Drug Interactive Graph.} The knowledge graph is a multi-relational semantic network~\cite{zhang2023application,zhao2024drug}. Endowed with rich domain information, the knowledge graph contains complex associations between biomedical entities~\cite{wang2017knowledge,wang2020gognn}. KGNN~\cite{lin2020kgnn} was the first to introduce GCN~\cite{kipf2017semisupervised} to encode the structured relationships in the knowledge graph. By spreading information in the graph and considering the relationships between entities simultaneously, it learned a comprehensive representation of drugs. MDNN~\cite{lyu2021mdnn} proposed a two-path framework to learn representations from multi-modal data and explore the similarities between modalities from multiple sources.
While these methods achieve promising results, they ignore the local molecular structures of drugs.

\begin{figure*}[!t]
    \centering
    \includegraphics[width=\linewidth]{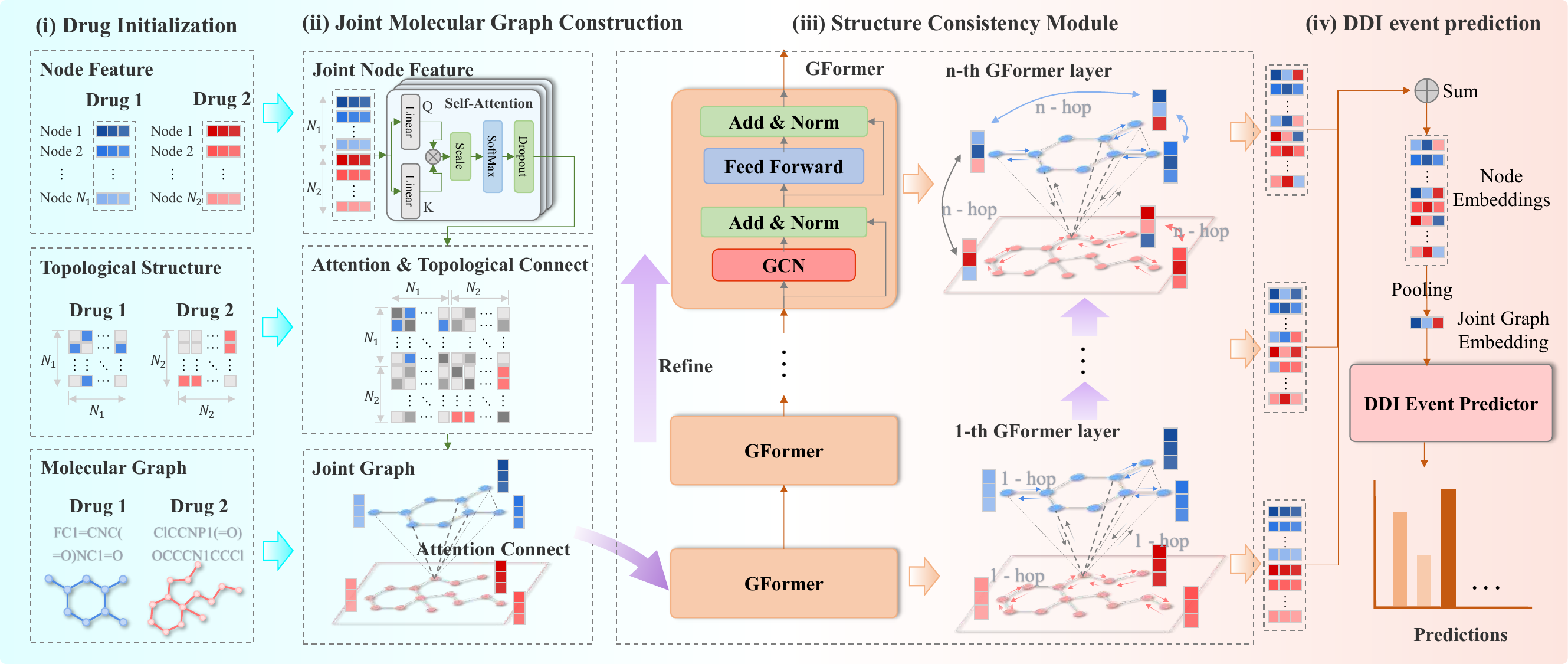}
    \caption{Framework of MolBridge. Given a pair of molecular SMILES as input, the model performs: (i) drug initialization to obtain node features and adjacency relationships for each molecule, (ii) joint molecular graph construction to capture inter-molecular connections and cross-drug interactions, (iii) structure consistency module to learn atomic dependencies within the joint graph, and (iv) DDI event prediction via a final classifier that outputs interaction probabilities.}
    \label{fig:Framework}
\end{figure*}

\noindent\textbf{Methods Based on Molecular Structure and Interactive Graph.}
MRCGNN\cite{xiong2023multi} encodes molecular graphs using TrimNet\cite{li2021trimnet} and refines drug representations via R-GCN~\cite{schlichtkrull2018modeling} within a drug association network. It also incorporates contrastive learning to enhance prediction. RaGSECo\cite{jiang2024relation} updates drug representations through DDI and drug similarity graphs, enabling effective learning for drugs with limited labeled data. TIGER\cite{su2024dual} introduces a dual-channel heterogeneous graph framework that integrates molecular structures and knowledge graph data, using a relation-aware self-attention mechanism to capture diverse interaction patterns.
In contrast,
MolBridge focuses purely on the chemical structures of drug pairs and explicitly models atom-level interactions and refines joint molecular graphs, enabling a deeper structural understanding of DDIs, especially for metabolism-driven events, without requiring external annotations.

\section{Methodology}
We first formulate the DDI event prediction task and then introduce the key components of the proposed architecture. The overall framework of MolBridge is shown in Figure~\ref{fig:Framework}.

\subsection{Problem Formulation}
We study multi-class drug–drug interaction (DDI) event prediction fora drug pair $(d_i, d_j)$,  where each drug is represented as a molecular graph. The goal is to estimate the probability of each interaction type given the two molecules. Formally, we denote the predictor as
\begin{align}
    \hat{y}(d_i, d_j) = \Gamma\!\left((d_i, d_j)\mid \Theta, \mathcal{F}\right),
\end{align}
where $\Theta$ are learnable parameters and $\mathcal{F}$ denotes the joint graph representation learning procedure operating on the input drug pair. The output $\hat{y}$ is a multi-class probability vector over DDI event types.
% We focus on predicting the potential drug interaction event between drugs. The prediction is achieved by modeling the atom-based intrinsic relations within molecules.
% Our DDI event prediction can be formulated as a multi-classification task, aiming to estimate the probability of corresponding interaction event. Specifically, given a pair of drugs $(d_i, d_j)$, we propose a model to identify the interaction event denoted as:
% $\hat{y}(d_i,d_j)=\Gamma((d_i,d_j)|\Theta,\mathcal{F})$,
% where $\Theta$ is the model parameters and $\mathcal{F}$ denotes the joint graph representation learning model for the input drug pairs.

\subsection{Overview}
MolBridge learns pairwise molecular representations by explicitly modeling atom-level relations across two drugs. As illustrated in Figure~\ref{fig:Framework}, the framework consists of four modules: (i) \textit{Drug Initialization}, which parses SMILES strings and encodes atom-level features; (ii) \textit{Joint Molecular Graph Construction}, which builds a unified atom graph for the drug pair and introduces cross-molecular interactions; (iii) \textit{Structure Consistency Module} (SCM), which refines node features while preserving structural fidelity and capturing long-range dependencies; and (iv) \textit{DDI Event Classifier}, which predicts the interaction probabilities from aggregated node features.
% MolBridge aims to learn the essential representations of input drugs (i.e., a pair of molecules) by enhancing the atom-based relations between them. Specifically, MolBridge proposes a graph joint learning framework with a residual mechanism to represent a pair of drugs. As illustrated in Figure~\ref{fig:Framework}, MolBridge comprises four modules: (i) the \textit{Drug Initialization} module initializes the molecular structure and informative features, (ii) the \textit{Joint molecular Graph Construction} module learns a joint atom-based structure between input drugs, (iii) the \textit{Structure Consistency Module} models the long-distance atom associations within joint graph, and (iv) the \textit{DDI event prediction} via a final classifier that outputs interaction probabilities. MolBridge can achieve accurate predictions based on these designed modules by mining the potential associations between the paired drugs.

\subsection{Drug Initialization}
We use Simplified Molecular Input Line Entry System (SMILES ~\cite{weininger1988smiles}) to represent the given drugs $d_i$ and $d_j$. Through RDKit~\cite{bento2020open}, we adopt various atomic descriptors and perform feature encoding on the drug molecule graphs to obtain the node feature matrices $F_i$ and $F_j$, where $F_i\in \mathbb{R}^{N_i\times d}$, $F_j\in \mathbb{R}^{N_j\times d}$, and $d$ is the dimension of atomic features, $N_i$ (resp., $N_j$) denotes the atom numbers of drug $d_i$ (resp., $d_j$). Besides, we acquire the adjacency matrices of drugs $d_i$ and $d_j$ as $A_i\in\mathbb{R}^{N_i\times N_i}$ and $A_j\in\mathbb{R}^{N_j\times N_j}$. We consider the $(F_i,A_i)$ and $(F_j,A_j)$ represent the featured graphs of input drugs $d_i$ and $d_j$, respectively. 
%\textcolor{red}{The details of node features F are provided in Supplementary Material \footnote{https://anonymous.4open.science/r/MolBridge-6561/Supplymentary} Section 1 (Table 1).}
% The details of node features $F$ refer to Table ~\ref{tab:atom_feature_app} in the Appendix.
\textcolor{black}{The details of node features F are provided in Table 1 of Supplementary Material\footnote{https://anonymous.4open.science/r/MolBridge-6561/Supplymentary}.}

\subsection{Joint Molecular Graph Construction}
To extract the potential associations between molecules for DDI event prediction, we design a joint molecular graph construction module to model atom-level interactions between drug pairs.

\noindent\textbf{Joint Graph Formation.}
Given two individual drug molecular graphs $(F_i,A_i)$ and $(F_j,A_j)$, we jointly construct them into a unified graph $(F', A')$ as follows:
\begin{align}
    F'=\begin{bmatrix}F_i\\F_j\end{bmatrix},
    \hspace{1cm}
    A'=\begin{bmatrix}A_i & 0\\0 & A_j\end{bmatrix},
\end{align}
where $F'\in \mathbb{R}^{(N_i+N_j)\times d}$ is the concatenated node features and $A'\in \mathbb{R}^{(N_i+N_j)\times (N_i+N_j)}$ represents the joint adjacency matrix with block-diagonal structure preserving intra-molecular chemical bonds.

\noindent\textbf{Feature Projection.}
To obtain the concatenated features for subsequent graph processing, we apply a linear transformation to map them to the target dimensional space:
\begin{align}
    H = \text{MLP}(F'),
\end{align}
where $H\in \mathbb{R}^{(N_i+N_j)\times dim}$ represents the projected joint node features, with $dim$ being the hidden dimension for downstream processing.

\noindent\textbf{Cross-Molecular Interaction Modeling.}
With the joint graph representation, we employ multi-head self-attention to capture implicit associations between atoms across the two molecules:
\begin{align}
    A_r = \text{MultiHeadAttention}(H),
\end{align}
where $A_r\in \mathbb{R}^{(N_i+N_j)\times (N_i+N_j)}$ captures the learned cross-molecular atom interactions, complementing the explicit chemical bond structure in $A'$.

\noindent\textbf{Graph Structure Integration.}
To adaptively balance the influences between intrinsic molecular structures $A'$ and learned atom interactions $A_r$, we integrate the explicit them:
\begin{align}
    A = (1 - \alpha)A'+\alpha A_r,
\end{align}
where $\alpha$ is a learnable parameter that adaptively balances the contribution of explicit chemical bonds and implicit cross-molecular interactions.

\subsection{Structure Consistency Module}
To preserve the structural fidelity of the original molecules and enhance the quality of learned atomic interactions, we introduce a novel Structure Consistency Module (SCM). SCM enforces alignment between the input molecular graph $(H, A)$ and the learned representations, while effectively capturing long-range atomic dependencies.

\noindent\textbf{Multi-layer Graph Processing.}
The SCM consists of multiple stacked GFormer layers, each designed to refine node representations through neighborhood aggregation and feature transformation:
\begin{align}
    & F^l = \text{GFormer}(F^{l - 1}, A),
\end{align}
where $F^l$ denotes the node features after the $l$-th GFormer layer, with $F^0 = H$ as the initial features.

\noindent\textbf{GFormer Layer Architecture.}
Each GFormer layer employs a sequential architecture that \textcolor{black}{combines graph convolution $X^l$ with feed-forward transformation $F^l$ as follows}:
\begin{align}
    & X^l = \text{LayerNorm}(\text{GCN}(F^{l-1}, A)) + F^{l-1}, \\
    & F^l = \text{LayerNorm}(\text{FFN}(X^l) + X^l),
\end{align}
where \textcolor{black}{$X^l$ represents the intermediate node features after graph convolution and residual connection in the $l$-th layer and} \textcolor{black}{GCN}~\cite{kipf2017semisupervised} performs graph convolution for neighborhood aggregation:
\begin{align}
    \text{GCN}(F^{l-1}, A) = (A+I)F^{l-1},
\end{align}
\textcolor{black}{where $I \in \mathbb{R}^{(N_i+N_j) \times (N_i+N_j)}$ is the identity matrix that enables self-loops for each node, }and FFN represents the feed-forward network:
\begin{align}
    \text{FFN}(X^l) = \sigma(X^lW_1^l + B_1^l)W_2^l + B_2^l,
\end{align}
where $\sigma$ is the activation function, and $W_1^l$, $W_2^l$, $B_1^l$, $B_2^l$ are learnable parameters of the $l$-th layer.

\noindent\textbf{Multi-scale Feature Aggregation.}
To preserve information from different abstraction levels and enhance model robustness, we aggregate features from all GFormer layers:
\begin{align}
    \textbf{h} = \sum_{l=0}^n \sum_{i=1}^{N_1+N_2} F_i^l,
\end{align}
where $F_i^l$ represents the feature vector of the $i$-th atom from the $l$-th layer, and $n$ is the total number of GFormer layers. This aggregation strategy allows the model to capture both local structural patterns and long-range dependencies across the joint molecular graph.

\noindent\textbf{Design Rationale of SCM.}
The SCM is designed to tackle two fundamental challenges in molecular graph representation: (1) capturing long-range atomic interactions essential for modeling complex drug–drug interaction mechanisms, and (2) alleviating the over-smoothing problem that hinders graph-based molecular representation learning. To this end, SCM integrates residual connections and layer normalization within each GFormer layer to preserve gradient flow and feature distinctiveness. Additionally, its multi-scale aggregation strategy enables the retention of structural information across different levels of molecular abstraction.

\begin{table*}
\centering
\caption{Results of MolBridge and baselines for DDI event prediction on two datasets.}
\begin{tabular}{lcccccccc}
\toprule
\multirow{2}{*}{Method} & \multicolumn{4}{c}{\textbf{Deng Dataset}} & \multicolumn{4}{c}{\textbf{Ryu Dataset}} \\
\cmidrule(lr){2-5} \cmidrule(lr){6-9}
                    & Acc.   & Macro-F1 & Macro-Pre. & Macro-Rec. & Acc.   & Macro-F1 & Macro-Pre. & Macro-Rec. \\
\midrule
DeepDDI             & 78.07  & 60.55    & 66.11      & 58.39      & 93.23  & 86.43    & 89.28      & 85.12      \\
R-GCN               & 86.95  & 70.26    & 75.00      & 68.78      & 92.84  & 84.87    & 88.81      & 82.91      \\
TrimNet-DDI         & 85.70  & 65.48    & 70.46      & 63.63      & 93.53  & 82.88    & 86.27      & 81.28      \\
GoGNN               & 87.66  & 69.38    & 73.16      & 68.41      & 94.24  & 85.89    & 89.49      & 84.51      \\
SSI-DDI             & 78.66  & 42.16    & 51.39      & 38.96      & 90.08  & 66.63    & 75.07      & 62.87      \\
MUFFIN              & 82.69  & 52.45    & 62.04      & 48.44      & 95.10  & 85.66    & 89.80      & 83.39      \\
MRCGNN              & \underline{89.79} & \underline{77.91} & \underline{81.01} & \underline{76.88} & \textbf{95.67} & \underline{88.94} & \underline{92.21} & \underline{87.27} \\
DSN-DDI             & 84.02  & 63.19    & 69.16      & 60.42      & 94.58  & 84.79    & 89.45      & 82.42      \\
CSSE-DDI            & 82.90  & 63.46    & 70.05      & 61.19      & 90.90  & 87.21    & 89.82      & 85.64      \\
TIGER               & 87.21  & 70.58    & 73.20      & 70.20      & 93.40  & 84.21    & 86.73      & 83.52      \\
\multirow{2}{*}[-0.0em]{\textbf{MolBridge}} & \textbf{92.39} & \textbf{84.83} & \textbf{87.45} & \textbf{84.00} & \underline{95.40} & \textbf{91.13} & \textbf{93.29} & \textbf{89.76} \\
 & (±0.28) & (±1.61) & (±1.84) & (±1.96) & (±0.14) & (±1.01) & (±0.90) & (±1.28) \\
\midrule
\textit{Improv.} (\%) & 2.6    & 6.92     & 6.44       & 7.12       & -0.27  & 2.19     & 1.08       & 2.49       \\
\bottomrule
\end{tabular}
\label{tab:comparison}
\end{table*}

\subsection{Model Training}
Given a pair of drugs $d_i$ and $d_j$, we use the joint graph molecular construction module and Structure Consistency Module to encode the atom-level interactions between them as the overall representation $\textbf{h}$. The final prediction is obtained through:
\begin{align}
    \hat{y}(d_i, d_j) = \text{SoftMax}(\text{MLP}(\textbf{h})),
\end{align}
where MLP represents a multi-layer perceptron that maps the joint graph representation to the DDI event probability distribution.

\paragraph{Loss Function.}
For the multi-class DDI event prediction task, we employ the cross-entropy loss function:
\begin{align}
    \mathcal{L} = -\sum_{c = 1}^{C}y^{c}(d_i,d_j)\log(\hat{y}^{c}(d_i,d_j)),
\end{align}
where \textit{C} is the number of DDI event classes, $y^c(di, dj)$ is the ground truth label (1 if the sample belongs to class c, 0 otherwise), and $\hat{y}^c(di, dj)$ represents the predicted probability for class $c$.

\paragraph{Training Process.}
The model is trained end-to-end using gradient-based optimization. During training, the joint molecular graph construction and Structure Consistency Module are optimized to minimize the prediction loss, enabling the model to learn both explicit chemical bond patterns and implicit inter-molecular associations that are critically relevant for accurate DDI event prediction.

\subsection{Computational Complexity Analysis}

\textcolor{black}{
 We analyze the time complexity of MolBridge for a joint graph with $N = N_i + N_j$ atoms, where each node has hidden dimensionality $dim$ after projection, the FFN in each layer uses hidden size $d_{\mathrm{hid}}$, and the Structure Consistency Module has $L$ stacked GFormer layers. Cross- molecular interaction modeling via multi-head self-attention over joint node features costs $\mathcal{O}(N^2 \cdot dim)$ per forward pass. Each GFormer layer applies dense graph propagation $(A{+}I)F$ followed by an FFN, incurring $\mathcal{O}(N^2 \cdot dim + N \cdot dim \cdot d_{\mathrm{hid}})$. Across $L$ layers, the Structure Consistency Module thus requires $\mathcal{O}\big(L \cdot (N^2 \cdot dim + N \cdot dim \cdot d_{\mathrm{hid}})\big)$. Therefore, the end-to-end time complexity is $\mathcal{O}\big(N^2 \cdot dim + L \cdot (N^2 \cdot dim + N \cdot dim \cdot d_{\mathrm{hid}})\big)$, typically dominated by the quadratic terms in $N$. In practice, we cap atoms at 50 per drug (i.e., $N \le 100$), which keeps these costs tractable in our experiments.
}

\section{Experiments}

\subsection{Experimental Settings}

\noindent\textbf{Dataset.} We use two benchmarking datasets to assess the performance of MolBridge.
Specifically, Deng's dataset~\cite{deng2020multimodal} encompasses a total of 37,264 samples involving 579 distinct drugs with 65 types of DDI event. The Ryu's dataset~\cite{ryu2018deep} consists of 191,570 samples involving 1,700 different drugs with 86 types of DDI event. Each sample within these datasets is represented as a tuple \((\text{d}_1, \text{d}_2, c)\), where \(\text{d}_1\) and \(\text{d}_2\) denote the SMILES sequences of \(\text{drug}_1\) and \(\text{drug}_2\), respectively. And \(c\) designates the interaction type.

\noindent\textbf{Baselines.} \textcolor{black}{We compare MolBridge against representative baselines across three experimental settings. For the main DDI event prediction task (Table~\ref{tab:comparison}) and frequency-stratified evaluation (Table~\ref{tab:freq_combined}), we select: DeepDDI~\cite{ryu2018deep}, R-GCN~\cite{schlichtkrull2018modeling}, TrimNet-DDI~\cite{li2021trimnet}, GoGNN~\cite{wang2020gognn}, SSI-DDI~\cite{nyamabo2021ssi}, MUFFIN~\cite{10.1093/bioinformatics/btab169}, MRCGNN~\cite{xiong2023multi}, DSN-DDI~\cite{li2023dsn}, CSSE-DDI~\cite{du2024customized}, and TIGER~\cite{su2024dual}. For the inductive generalization evaluation on DrugBank (Table~\ref{tab:inductive_merged}), we compare against methods with reported cold-start performance, including: CSMDD~\cite{liu2022predict}, HIN-DDI~\cite{tanvir2023predicting}, KG-DDI~\cite{lin2020kgnn}, CompGCN~\cite{vashishth2020composition}, Decagon~\cite{zitnik2018modeling}, KGNN~\cite{lin2020kgnn}, and DeepLGF~\cite{ren2022deeplgf}. Note that methods marked with $^*$ leverage external knowledge graphs or heterogeneous networks, while MolBridge relies solely on molecular structures.}

% \noindent\textbf{Baselines.} We compare MolBridge against representative baselines from two main categories. The first category includes molecular structure-based approaches: DeepDDI~\cite{ryu2018deep}, SSI-DDI~\cite{nyamabo2021ssi}, and GoGNN~\cite{wang2020gognn}. The second category includes drug interaction event network-based approaches: R-GCN~\cite{schlichtkrull2018modeling}, TrimNet-DDI~\cite{li2021trimnet}, MRCGNN~\cite{xiong2023multi}, TIGRE~\cite{su2024dual}, and CSSE-DDI~\cite{du2024customized}.

\noindent\textbf{Implementation Details.}
\textcolor{black}{For the transductive experiments on Deng's and Ryu's datasets, we compare MolBridge with DeepDDI, R-GCN, GoGNN, TrimNet, SSI-DDI, MUFFIN, and MRCGNN using results reported in~\cite{xiong2023multi}, and re-implement DSN-DDI, CSSE-DDI, and TIGER based on their official configurations. Following the protocol in MRCGNN~\cite{xiong2023multi}, we perform five-fold cross-validation with a 7:1:2 train/validation/test split and evaluate using Accuracy, Macro-F1, Macro-Recall, and Macro-Precision metrics. For the inductive generalization experiments on DrugBank, we follow the evaluation protocol and data split settings in~\cite{zhang2023emerging}, and compare with baseline methods whose results are obtained from~\cite{zhang2023emerging}. During training, we set the batch size to 512, learning rate to 0.005, random seed to 42, and use 3-layer GFormer modules. We adopt AdamW as the optimizer and select the best-performing model based on Accuracy and Macro-F1 on the validation set (max epochs: 500). Final results are obtained by averaging performance across all folds. All experiments are conducted on a machine with an Intel Xeon Platinum-8457C CPU and an NVIDIA L20 (48GB) GPU. The software environment includes Ubuntu 22.04.1 LTS, PyTorch 2.4.0 (CUDA 11.8), and RDKit 2024.9.6.}
% Following the protocol in MRCGNN~\cite{xiong2023multi}, we perform five-fold cross-validation with a 7:1:2 train/validation/test split. We evaluate using Accuracy, Macro-F1, Macro-Recall, and Macro-Precision metrics, and report averaged results across all folds. 
% \textcolor{red}{Detailed experimental configurations and hardware specifications are provided in GitHub.}
% Detailed experimental configurations and hardware specifications are provided in the Appendix.

\begin{table*}[t]
  \centering
  \caption{Performance on the Deng test set stratified by label frequency. Left: high-frequency labels (0--3). Right: rare labels (35--64). MolBridge consistently surpasses baselines, with particularly strong gains on rare classes.}
  \begin{tabular}{l*{8}{c}}
  \toprule
  \multirow{2}{*}{\textbf{Methods}} & \multicolumn{4}{c}{\textbf{High-frequency labels (0--3)}} & \multicolumn{4}{c}{\textbf{Rare labels (35--64)}} \\
  \cmidrule(lr){2-5} \cmidrule(lr){6-9}
  & \textbf{Acc.} & \textbf{Macro-F1} & \textbf{Macro-Prec.} & \textbf{Macro-Rec.} & \textbf{Acc.} & \textbf{Macro-F1} & \textbf{Macro-Prec.} & \textbf{Macro-Rec.} \\
  \midrule
  DSN-DDI & 85.77±0.51 & 87.03±0.49 & 89.92±0.25 & 84.43±0.70 & 50.00±4.44 & 43.68±4.78 & 57.59±4.97 & 37.53±4.13 \\
  TIGER & \underline{88.30±0.35} & \underline{89.57±0.38} & \underline{92.14±0.21} & \underline{87.20±0.52} & \underline{66.76±3.81} & \underline{62.06±5.81} & \underline{73.36±6.83} & \underline{56.56±5.35} \\
  \textbf{MolBridge} & \textbf{93.17±0.38} & \textbf{93.63±0.44} & \textbf{95.18±0.38} & \textbf{92.16±0.58} & \textbf{82.03±3.59} & \textbf{81.74±4.23} & \textbf{89.44±4.58} & \textbf{77.53±4.41} \\
  \midrule
  \textit{Improv.} (\%) & \textbf{+5.5} & \textbf{+4.5} & \textbf{+3.3} & \textbf{+5.7} & \textbf{+22.9} & \textbf{+31.7} & \textbf{+21.9} & \textbf{+37.1} \\
  \bottomrule
  \end{tabular}
  \label{tab:freq_combined}
\end{table*}

\subsection{Main Results}
\noindent\textbf{Performance Comparision.}
Table ~\ref{tab:comparison} presents the comparison performance of MolBridge and baselines on Deng's and Ryu's datasets, where the best and the second-best results are shown in bold and underlined, respectively. As shown in Table ~\ref{tab:comparison}, it can be observed that MolBridge outperforms the other eight baseline methods in DDI event prediction across two datasets. This demonstrates its effectiveness in predicting DDI event. We also have the following observations: (1) Compared with DeepDDI, TrimNet-DDI, and SSI-DDI, which only consider the local structural information of the drug molecule, MolBridge makes improvements of 25.61\%, 20.37\%, and 45.04\% on Deng's dataset, as well as 4.64\%, 8.48\%, and 26,89\% on Ryu's dataset in terms of Macro-Rec, which obviously indicates the atom-based long-distance interactions learned by MolBridge benefit DDI event prediction. (2) Compared with R-GCN and MUFFIN, which only adopt drug interactive information, MolBridge surpasses them at least by 15.22\% on Deng's dataset, and 6.32\% on Ryu's dataset in terms of Macro-Rec, which implies that considering the local structural information from drug molecular graph is also beneficial to DDI event prediction. (3) Among all baselines, MRCGNN exhibits better performance than others, implying the advantages of integrating the learned features from the molecular graph and the interaction graph. Our proposed MolBridge still makes an average improvement of 1.17\% in Acc, 4.56\% in Macro-F1, 3.76\% in Macro-Pre, and 4.81\% in Macro-Rec. It may be attributed to the reason that MolBridge not only effectively integrates the local structural information and drug interaction information as MRCGNN does, but also captures the long-distance associations within the constructed joint graph by the proposed SCM.

\begin{figure}[t]
      \centering
      \includegraphics[width=\columnwidth]{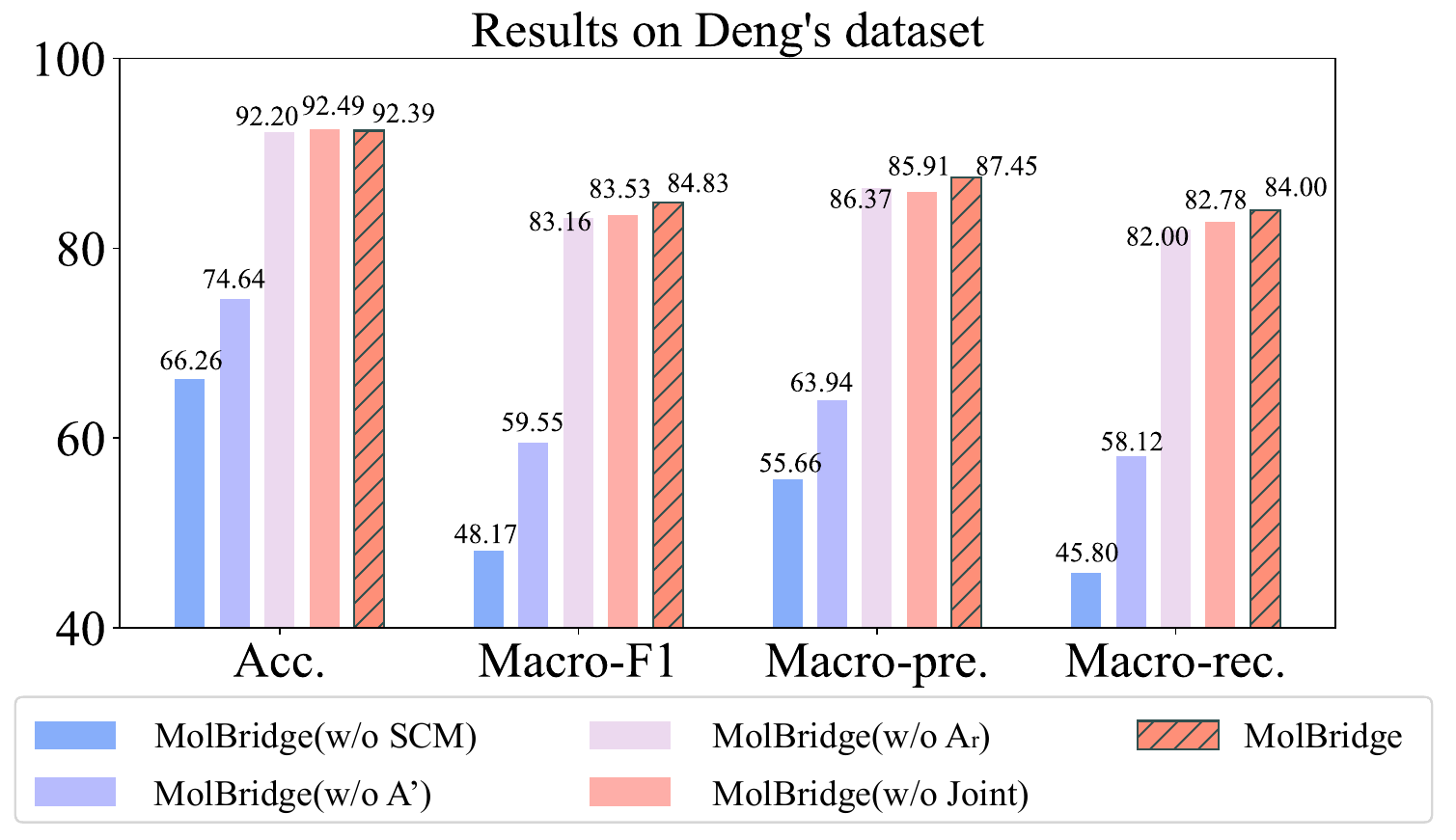}
      \caption{Ablation study results of MolBridge and its variants on Deng's dataset. The Structure Consistency Module (SCM) is the most critical component, while both $A'$ (explicit chemical bonds) and $A_r$ (attention-based cross-molecular interactions) contribute to optimal performance.}
      \label{fig:ablation_main}
  \end{figure}

\noindent\textbf{Ablation Study.}
To investigate the contribution of different components in our proposed MolBridge model, we conduct a comprehensive ablation study on Deng's dataset. We consider four variants: (1) \textbf{w/o SCM}: replacing the structure consistency module with standard GCN layers, (2) \textbf{w/o $A'$}: removing the original chemical bond adjacency matrix and using only the reconstructed graph, (3) \textbf{w/o $A_r$}: eliminating the attention-based graph reconstruction and using only explicit chemical bonds, and (4) \textbf{w/o Joint}: removing the joint construction strategy and modeling drugs independently before fusion.
\textcolor{black}{As shown in Figure~\ref{fig:ablation_main},} the results demonstrate that all components contribute significantly to DDI event prediction performance, with the Structure Consistency Module being the most critical component (accuracy drops from 92.39\% to 66.26\% without SCM). The joint molecular graph construction strategy also proves essential, as independent drug modeling leads to substantial performance degradation. Notably, both explicit chemical bonds ($A'$) and reconstructed inter-molecular associations ($A_r$) are necessary, with their combination achieving optimal performance through complementary structural information. 
% \textcolor{red}{Full ablation results are provided in GitHub.}
% Detailed results are provided in the Appendix Figure~\ref{fig:ablation_app}.

\noindent\textbf{Hyperparameter Sensitivity Analysis.}
We analyze the sensitivity of MolBridge by varying key hyperparameters including learning rate, batch size, attention heads, and SCM layers in Deng's dataset. The optimal configuration achieves peak performance at learning rate 0.005, batch size 512, 4 attention heads, and 3 SCM layers. 
\textcolor{black}{Complete hyperparameter sensitivity curves and settings are provided in Supplementary Material Section 3 (Figures 1-4).}
% The complete details of the hyperparameter sensitivity analysis are provided in Appendix Figure~\ref{fig:hparam_app}.

\subsection{Frequency-Stratified Evaluation}
\textcolor{black}{To assess robustness under class imbalance, we evaluate MolBridge and baselines on the Deng dataset by stratifying the test set into high-frequency (labels 0--3) and rare (labels 35--64) DDI types. The training distribution is highly skewed (on average 19{,}136 vs. 481 samples per fold for frequent vs. rare labels). As summarized in Table~\ref{tab:freq_combined}, MolBridge surpasses TIGER on frequent labels by 5.5\% Accuracy and 4.5\% Macro-F1, and delivers substantially larger gains on rare labels (+22.9\% Accuracy, +31.7\% Macro-F1, +21.9\% Macro-Precision, +37.1\% Macro-Recall). These results indicate that joint molecular graph construction together with the structure consistency module yields discriminative and robust representations under skewed label distributions.}

 \begin{table}[t]
    \centering
    \caption{Inductive evaluation on DrugBank under S1 (one seen drug, one unseen) and S2 (two unseen). Methods
  marked with $^*$ use external knowledge graphs or heterogeneous networks.}
    \begin{tabular}{lcccc}
    \toprule
    \multirow{2}{*}{\textbf{Methods}} & \multicolumn{2}{c}{\textbf{S1}} & \multicolumn{2}{c}{\textbf{S2}} \\
    \cmidrule(lr){2-3}\cmidrule(lr){4-5}
    & \textbf{Macro-F1} & \textbf{Acc} & \textbf{Macro-F1} & \textbf{Acc} \\
    \midrule
    CSMDD & \underline{45.5 ± 1.8} & \underline{62.6 ± 2.8} & \underline{19.8 ± 3.1} & \underline{37.3 ± 4.8} \\
    HIN-DDI$^*$ & 37.3 ± 2.9 & 58.9 ± 1.4 & 8.8 ± 1.0 & 27.6 ± 2.4 \\
    KG-DDI$^*$ & 26.1 ± 0.9 & 46.7 ± 1.9 & 1.1 ± 0.1 & 32.2 ± 3.6 \\
    CompGCN$^*$ & 26.8 ± 2.2 & 48.7 ± 3.0 & -- & -- \\
    Decagon$^*$ & 24.3 ± 4.5 & 47.4 ± 4.9 & -- & -- \\
    KGNN$^*$ & 23.1 ± 3.4 & 51.4 ± 1.9 & -- & -- \\
    DeepLGF$^*$ & 39.7 ± 2.3 & 60.7 ± 2.4 & 4.8 ± 1.9 & 31.9 ± 3.7 \\
    \textbf{MolBridge} & \textbf{53.8 ± 5.0} & \textbf{64.4 ± 2.8} & \textbf{20.5 ± 1.3} & \textbf{40.3 ± 2.7} \\
    \bottomrule
    \end{tabular}
    \label{tab:inductive_merged}
  \end{table}

\begin{figure*}[!t]
    \centering
    \includegraphics[width=0.9\linewidth]{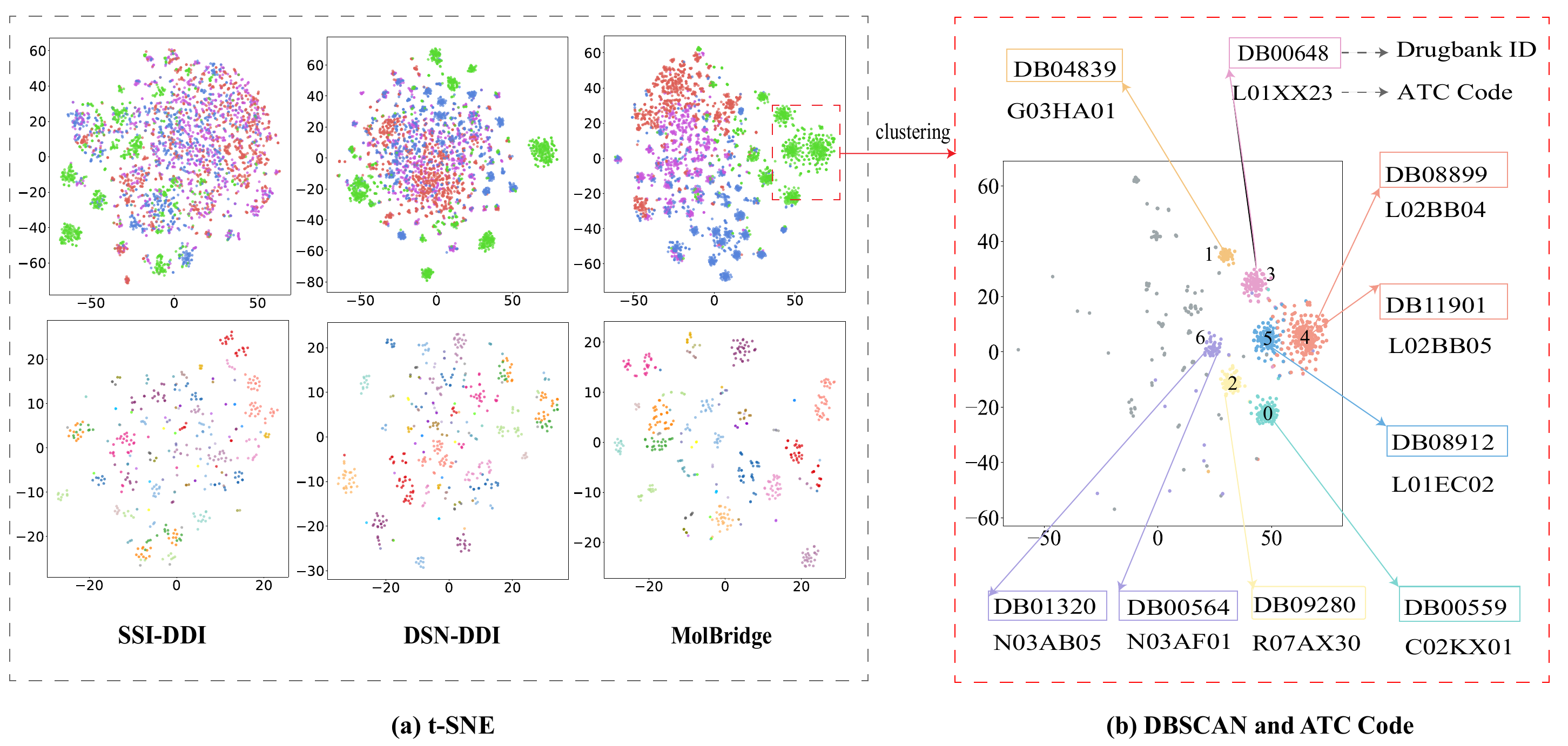}
    \caption{Visualization and cluster analysis of DDI representations. (a) t-SNE comparison of embedding quality across methods for high-frequency (upper) and low-frequency (lower) DDI event. (b) DBSCAN clustering identifies seven distinct sub-clusters within one interaction type, with drugs labeled by DrugBank IDs and ATC codes.}
    \label{fig:cluster}
\end{figure*}

\subsection{Inductive Generalization on DrugBank}
\textcolor{black}{We assess inductive generalization on DrugBank under S1 (one seen drug, one unseen) and S2 (both drugs unseen). Table~\ref{tab:inductive_merged} reports Macro-F1 and Accuracy; methods using external knowledge graphs or heterogeneous networks are marked with $^*$. MolBridge achieves the best results in both regimes, surpassing the strongest baseline (CSMDD) by +8.3 Macro-F1 / +1.8 Acc in S1 and by +0.7 Macro-F1 / +3.0 Acc in S2. These gains suggest that atom-level cross-molecular modeling with joint-graph refinement yields transferable representations. KG-augmented methods (e.g., DeepLGF$^*$) are competitive but still behind MolBridge in the strictly inductive S2 setting. Overall, all methods degrade from S1 to S2, while MolBridge maintains a consistent lead. }
% The complete results with all baselines and additional metrics (Kappa) are provided in Appendix Table~\ref{tab:inductive_complete_app}.

\subsection{Visualization and Cluster Analysis}

\begin{table}[t]
  \caption{ATC Classification for Clusters 4--6 (Levels 1--3). Abbr.: Antineopl.=Antineoplastic; immuno.=immunomodulating; Endocr.=Endocrine; ther.=therapy; Antiepi.=Antiepileptics; inh.=inhibitors.}
  \label{tab:atc_main}
  \centering
  \footnotesize
  \renewcommand{\arraystretch}{1.12}
  \setlength{\tabcolsep}{3.5pt}
  \begin{tabular}{ccp{0.8cm}p{1.75cm}p{1.55cm}p{1.45cm}}
  \toprule
  \textbf{Cls} & \textbf{Drug} & \textbf{ATC} & \textbf{L1} & \textbf{L2} & \textbf{L3} \\
  \midrule
  \multirow{2}{*}{4} & DB11901 & L02BB05 & L-Antineopl. immuno. & 02-Endocr. ther. & B-Hormones \\
  & DB08899 & L02BB04 & L-Antineopl. immuno. & 02-Endocr. ther. & B-Hormones \\
  \midrule
  5 & DB08912 & L01EC02 & L-Antineopl. immuno. & 01-Antineopl. & E-Kinase inh. \\
  \midrule
  \multirow{2}{*}{6} & DB01320 & N03AB05 & N-Nervous sys. & 03-Antiepi. & A-Antiepi. \\
  & DB00564 & N03AF01 & N-Nervous sys. & 03-Antiepi. & A-Antiepi. \\
  \bottomrule
  \end{tabular}
  \vspace{-0.3cm}
\end{table}

\noindent\textbf{t-SNE Visualization.} We employed t-SNE~\cite{van2008visualizing} to analyze embedding quality across SSI-DDI, DSN-DDI, and MolBridge, focusing on both high-frequency (top 4) and low-frequency (bottom 30) DDI events. As shown in Figure~\ref{fig:cluster}a, MolBridge produces notably more compact and well-separated clusters compared to baselines. Drug pairs with the same interaction type form tighter clusters in our method, indicating superior representation learning. This improvement is particularly pronounced for rare DDI events (lower row), where our method maintains clear clustering patterns while baselines show more scattered distributions, demonstrating effective integration of molecular graph information with the constructed joint graph. 
\textcolor{black}{Additional comparative visualizations with other baseline methods (MRCGNN-DDI and TrimNet-DDI) are provided in the Appendix for the low-frequency cases (Figure~\ref{fig:d-tsne-han}) and in the Supplementary Material Section 4 (Figure 5) for the frequent cases.}
% Additional comparative visualizations with other baseline methods (MRCGNN-DDI and TrimNet-DDI) are provided in Figures~\ref{fig:d-tsne-duo} and ~\ref{fig:d-tsne-han} in the Appendix.

\noindent\textbf{Fine-grained Cluster Analysis.} \textcolor{black}{We analyzed the internal structure within individual interaction types by extracting samples from one specific type (green points in Figure~\ref{fig:cluster}a). DBSCAN clustering identified seven distinct sub-clusters (Figure~\ref{fig:cluster}b), each exhibiting characteristic drug signatures: cluster 0 by DB00559 (ATC code C02KX01), cluster 1 by DB04839 (G03HA01), cluster 2 by DB09280 (R07AX30), cluster 3 by DB00648 (L01XX23), clusters 4--5 by DB08899/DB11901 (L02BB04/L02BB05) and DB08912 (L01EG02), and cluster 6 by DB01320/DB00564 (N03AB05/N03AF01). This demonstrates that our model captures meaningful distinctions within the same interaction category without explicit supervision.}

\noindent\textbf{Functional Similarity Analysis through ATC Codes.} \textcolor{black}{The spatial organization of sub-clusters reveals biologically meaningful patterns. Anatomical Therapeutic Chemical (ATC) classification analysis shows that representative drugs in clusters 3, 4, and 5 all belong to the "L" category (Antineoplastic and immunomodulating agents), explaining their spatial proximity in the embedding space. Similarly, the ATC code similarity between clusters 4 and 6 accounts for their close positioning. Remarkably, our model was trained exclusively on molecular graph structures without any therapeutic classification information, yet successfully organizes drugs according to their therapeutic similarities, demonstrating successful extraction of functional information from purely structural data. Table~\ref{tab:atc_main} shows the first 3 levels of ATC classification for clusters 4--6, with complete 5-level hierarchical information provided in Appendix Table~\ref{tab:atc_analysis_detailed}. Complete information for all clusters (0--6) is in Supplementary Material Section 5 (Table 2).}

\subsection{Structural interpretation for DDI mechanism}

\begin{figure}
    \centering
    \includegraphics[width=0.9\linewidth]{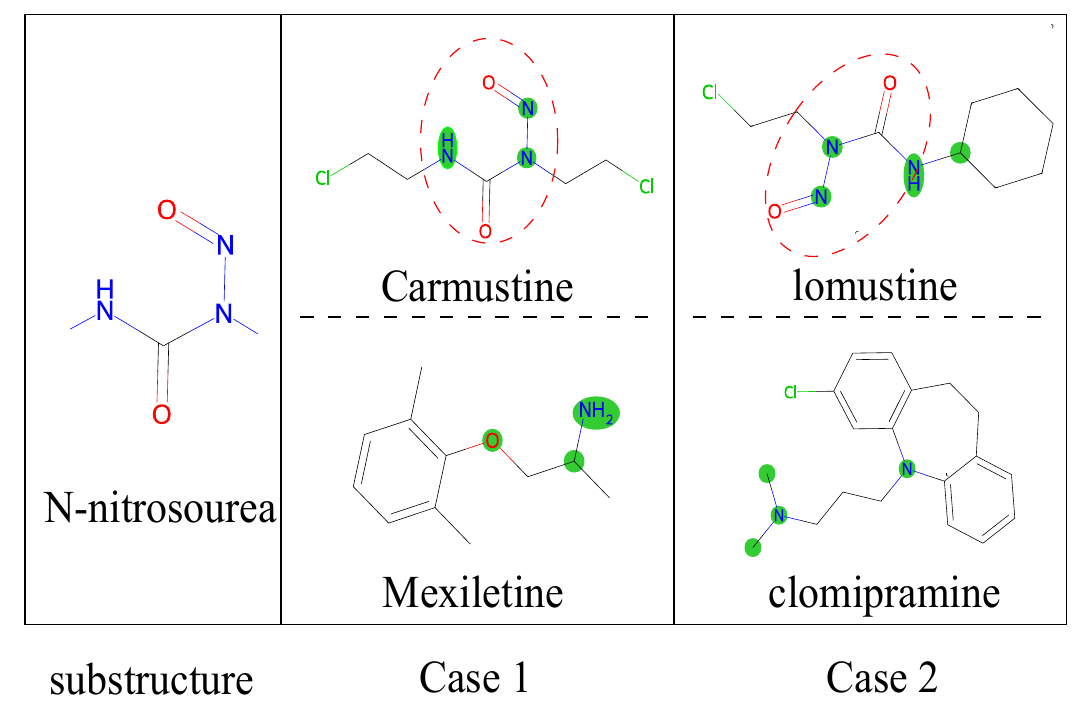}
    \caption{Representative substructures identified by MolBridge. The model highlights key atomic interactions (red box) linked to known reactive groups (e.g., N-nitrosourea), suggesting potential metabolic competition in DDI cases.}
    \label{fig:case}
\end{figure}

To validate MolBridge's effectiveness in identifying meaningful molecular interactions, we analyzed representative drug interaction cases with established pharmacological evidence. Figure~\ref{fig:case} shows the experimental results, where green-highlighted atoms represent nodes connected by higher-weight edges in the reconstructed joint graph, indicating critical interaction sites. Our model consistently identified the N-nitrosourea substructure as the key interaction component across all test cases, successfully highlighting N-nitrosourea groups in both Carmustine and Lomustine when paired with various drugs including Mexiletine, Bortezomib, Clomipramine, and Chlorpromazine. This finding is pharmacologically significant, as N-nitrosourea compounds are known alkylating agents that can lead to metabolic competition, enzyme inhibition, or formation of reactive intermediates, contributing to adverse drug-drug interactions~\cite{tew1983alkylating,fahrer2023dna}. The consistent identification of N-nitrosourea groups across different drug pairs validates MolBridge's ability to recognize genuine chemical interaction features, with comprehensive results provided in Appendix Figure~\ref{fig:comprehensive_cases}.

\subsection{Long-distance Interaction Analysis}

\begin{figure}
    \centering
    \includegraphics[width=\linewidth]{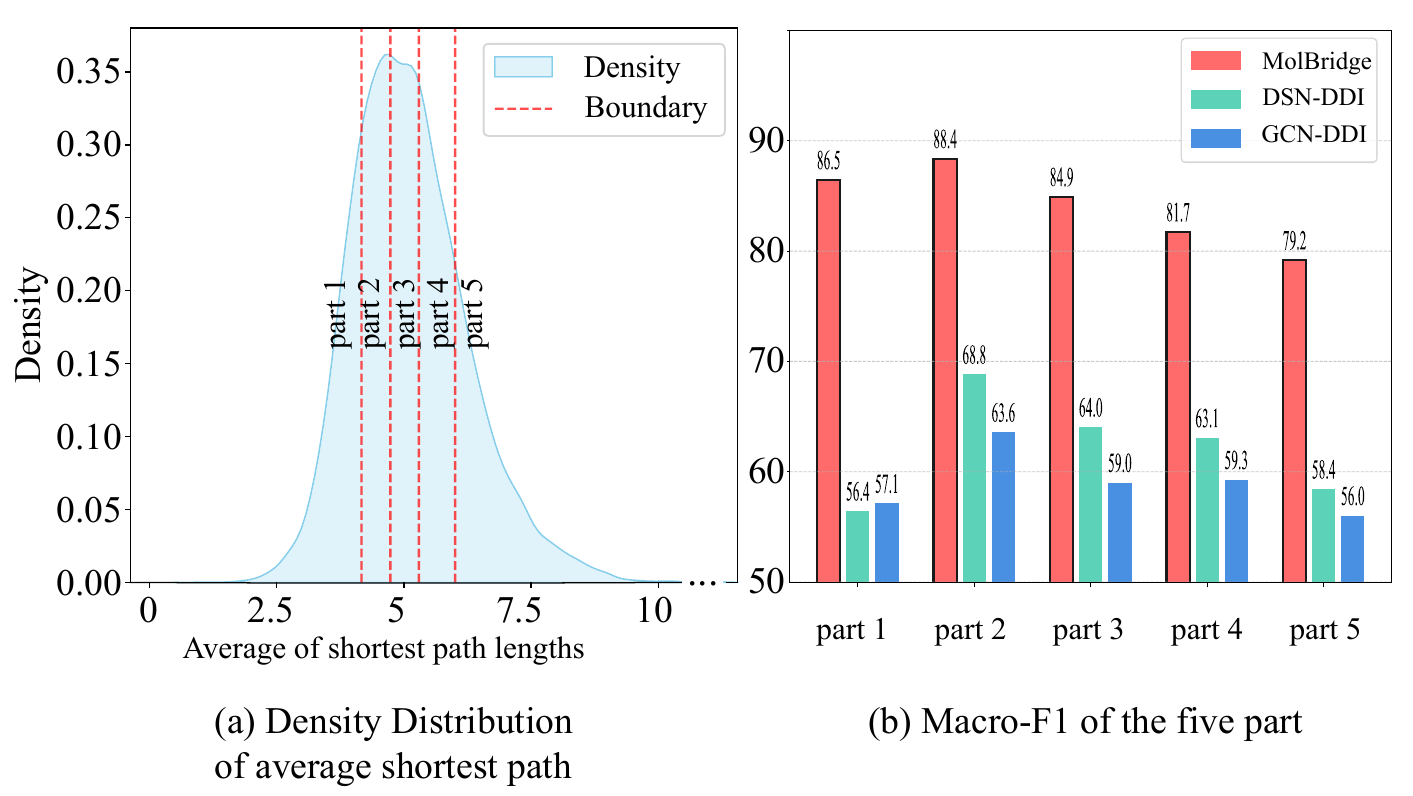}
    \caption{The analysis of long-range interaction on Deng's dataset. (a) Distribution of average shortest path lengths between atoms, with quintile markers. (b) Macro-F1 across distance groups, showing MolBridge's strength in modeling long-range interactions.}
    \label{fig:long_distance_compact}
\end{figure}

To evaluate the impact of long-distance atomic interactions on DDI event prediction, we stratified the Deng test set based on average shortest path lengths between atoms within drug molecules. Figure ~\ref{fig:long_distance_compact} shows the density distribution of these path lengths and the corresponding Macro-F1 performance comparison across five distance partitions.
The results demonstrate that MolBridge consistently outperforms baseline models across all distance ranges, with performance declining as inter-atomic distances increase. This validates our hypothesis that long-distance atomic interactions are critical for DDI event prediction accuracy. The details of the complete experimental results and analysis \textcolor{black}{are provided in Supplementary Material Section 6 (Figures 6-7).}

\section{Conclusion}
We propose MolBridge, a novel joint graph learning framework designed to improve DDI event prediction by holistically integrating intra-molecular structures with inter-molecular atom interactions. Unlike existing approaches that predominantly focus on either local structure patterns or isolated molecular representations, MolBridge introduces a Structure Consistency Module that effectively captures both localized chemical features and long-range dependencies across molecular graphs. This enables a more expressive and biochemically grounded representation of drug pairs, addressing key limitations of locality-restricted models. Extensive experiments on benchmark datasets demonstrate that MolBridge achieves state-of-the-art performance, outperforming several strong baselines in predicting complex pharmacological interactions. Looking forward, we plan to incorporate domain-specific pharmacological knowledge into the joint molecular graph construction, and further investigate motif-aware interaction modeling. 

%In future work, we will incorporate pharmacological knowledge into joint molecular graph construction and explore motif-based interaction modeling to further improve mechanistic interpretability and predictive accuracy.

\begin{acks}
We thank the anonymous reviewers for their valuable feedback and suggestions. This work was supported by research grants from [funding sources to be revealed after review].
\end{acks}

\bibliographystyle{ACM-Reference-Format}
\bibliography{ref}

\clearpage

\onecolumn

\begin{center}
    \Huge Appendix
\end{center}

\section{Over-smoothing Analysis}
\begin{figure}[!h]
    \centering
    \includegraphics[width=\columnwidth]{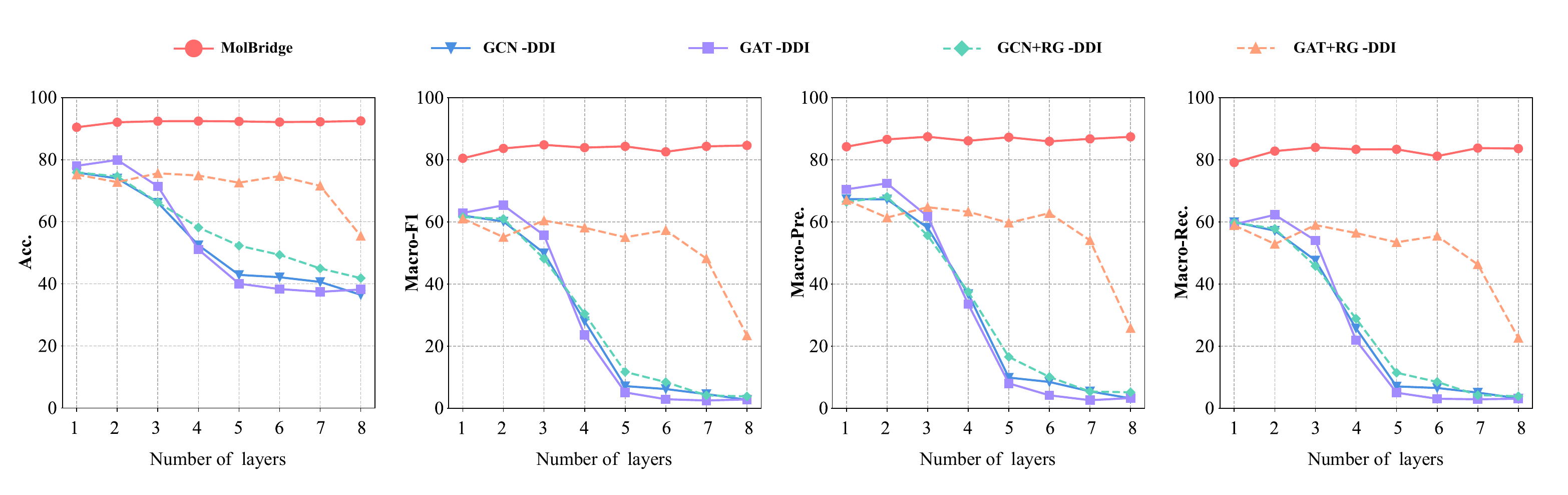}
    \caption{Experimental validation on Deng’s dataset of the over-smoothing in vanilla GNNs for DDI prediction. Figure illustrates the performance comparison between conventional GNNs (i.e., GCN and GAT) and our proposed MolBridge across varying network depths. The experimental results show that GCN and GAT exhibit performance deterioration as network depth increases, conclusively verifying the over-smoothing phenomenon specifically in the context of DDI prediction. While our proposed MolBridge maintains relatively stable performance across different depths, confirming its effectiveness in mitigating the over-smoothing in molecular interaction networks.}
    \label{fig:residul}
\end{figure}

We conducted comparison experiments between our proposed MolBridge and conventional GNNs on DDI event prediction across varying network depths. As illustrated in Figure \ref{fig:residul}, the experimental results provide clear evidence of the oversmoothing problem in DDI event prediction. Both GCN and GAT exhibit substantial performance decline as the number of layers increases beyond two, with score deteriorating dramatically with each additional layer. This empirical verification confirms that the over-smoothing significantly constrains the expressive capacity of traditional GNNs for DDI prediction, which is beneficial for capturing both nuanced interaction patterns and long-range dependencies across molecular structures.

\section{Detailed ATC Code Analysis.}

\textcolor{black}{The Anatomical Therapeutic Chemical (ATC) classification system divides drugs into hierarchical groups according to anatomical target, therapeutic usage, pharmacological mechanism, and chemical substance. We summarize representative ATC codes for the clusters most relevant to our main-text analysis (clusters 4–6) in Table~\ref{tab:atc_analysis_detailed}. The complete table covering all clusters (0–6) is provided in Supplementary Material Section 5 (Table 2).}

\begin{table*}[h]\footnotesize
  \caption{Representative ATC Code Analysis for Clusters 4--6}
  \label{tab:atc_analysis_detailed}
  \centering
  \renewcommand{\arraystretch}{1.3}
  \begin{tabular}{c|c|c|>{\raggedright\arraybackslash}p{2.2cm}|>{\raggedright\arraybackslash}p{2.2cm}|>{\raggedright\arraybackslash}p{2.2cm}|>{\raggedright\arraybackslash}p{2.2cm}|>{\raggedright\arraybackslash}p{2.0cm}}
  \toprule
  Cluster & Drug & ATC Code & Level 1 & Level 2 & Level 3 & Level 4 & Level 5 \\
  \midrule
  \multirow{2}{*}{4} & DB11901 & L02BB05 & L - Antineoplastic and immunomodulating agents & 02 - Endocrine therapy & B - Hormones and related agents & B - Antiandrogens & 05 - Enzalutamide \\
  & DB08899 & L02BB04 & L - Antineoplastic and immunomodulating agents & 02 - Endocrine therapy & B - Hormones and related agents & B - Antiandrogens & 04 - Enzalutamide \\
  \midrule
  5 & DB08912 & L01EC02 & L - Antineoplastic and immunomodulating agents & 01 - Antineoplastic agents & E - Protein kinase inhibitors & C - BRAF serine/ threonine kinase inhibitors & 02 - Dabrafenib \\
  \midrule
  \multirow{2}{*}{6} & DB01320 & N03AB05 & N - Nervous system & 03 - Antiepileptics & A - Antiepileptics & B - Hydantoin derivatives & 05 - Phenytoin \\
  & DB00564 & N03AF01 & N - Nervous system & 03 - Antiepileptics & A - Antiepileptics & F - Carboxamide derivatives & 01 - Carbamazepine \\
  \bottomrule
  \end{tabular}
\end{table*}

\section{Detailed t-SNE Visualization.}

\begin{figure*}[h]
    \centering
    \subfloat[TrimNet-DDI]{%
        \begin{minipage}[b]{0.2\linewidth}
            \includegraphics[width=\linewidth]{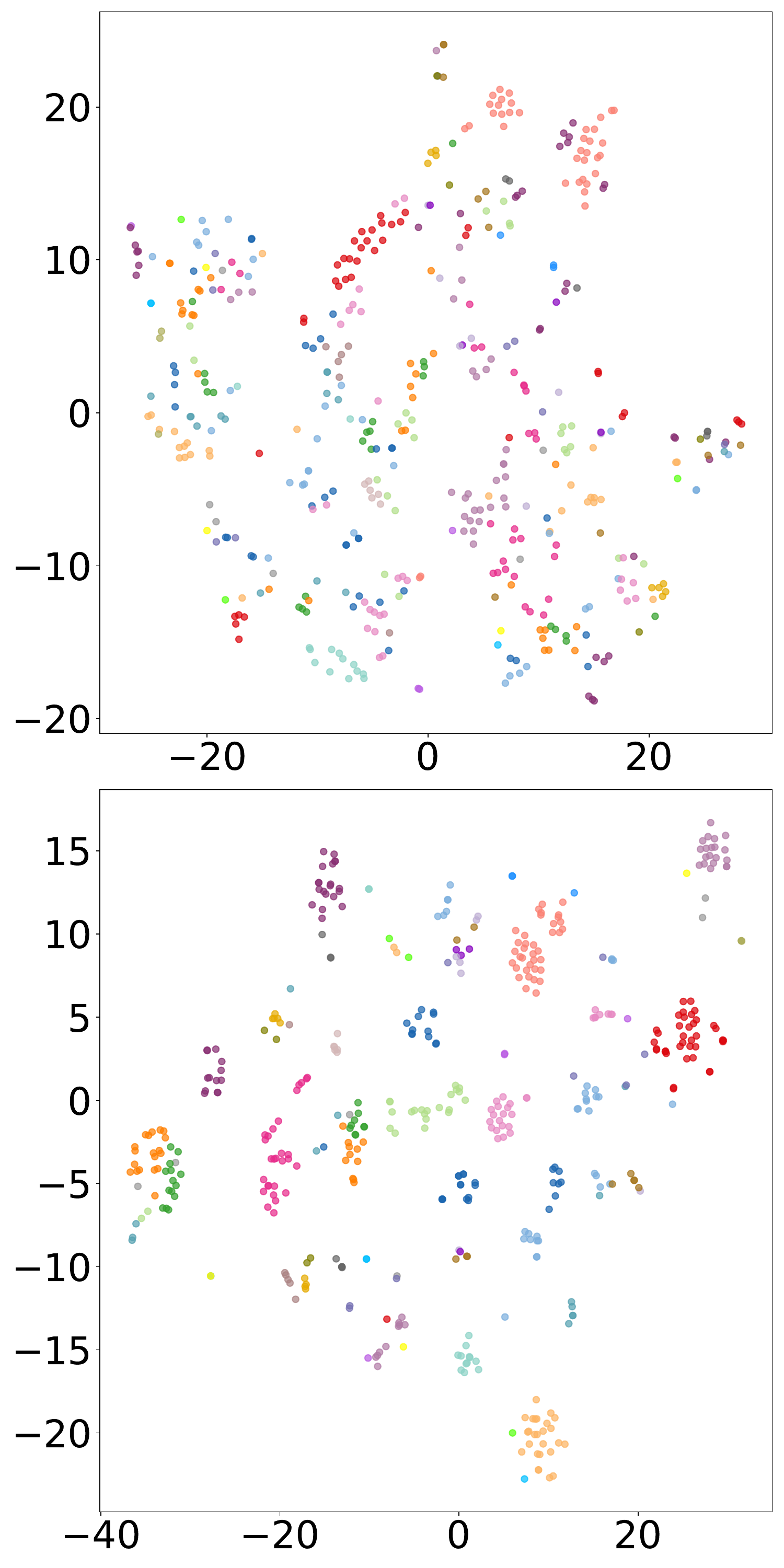}
        \end{minipage}%
    }\hfill
    \subfloat[SSI-DDI]{%
        \begin{minipage}[b]{0.2\linewidth}
            \includegraphics[width=\linewidth]{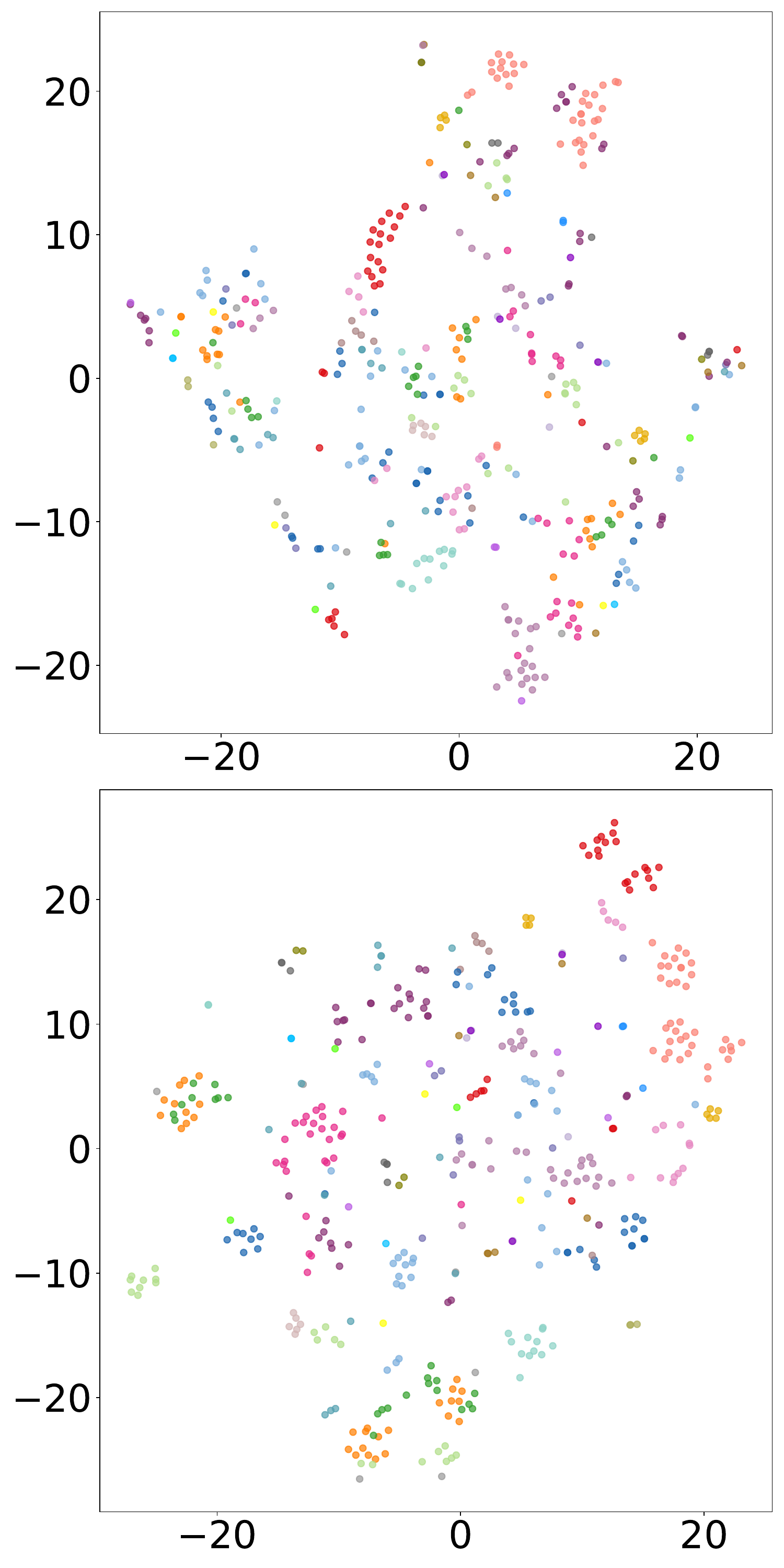}
        \end{minipage}%
    }\hfill
    \subfloat[MRCGNN]{%
        \begin{minipage}[b]{0.2\linewidth}
            \includegraphics[width=\linewidth]{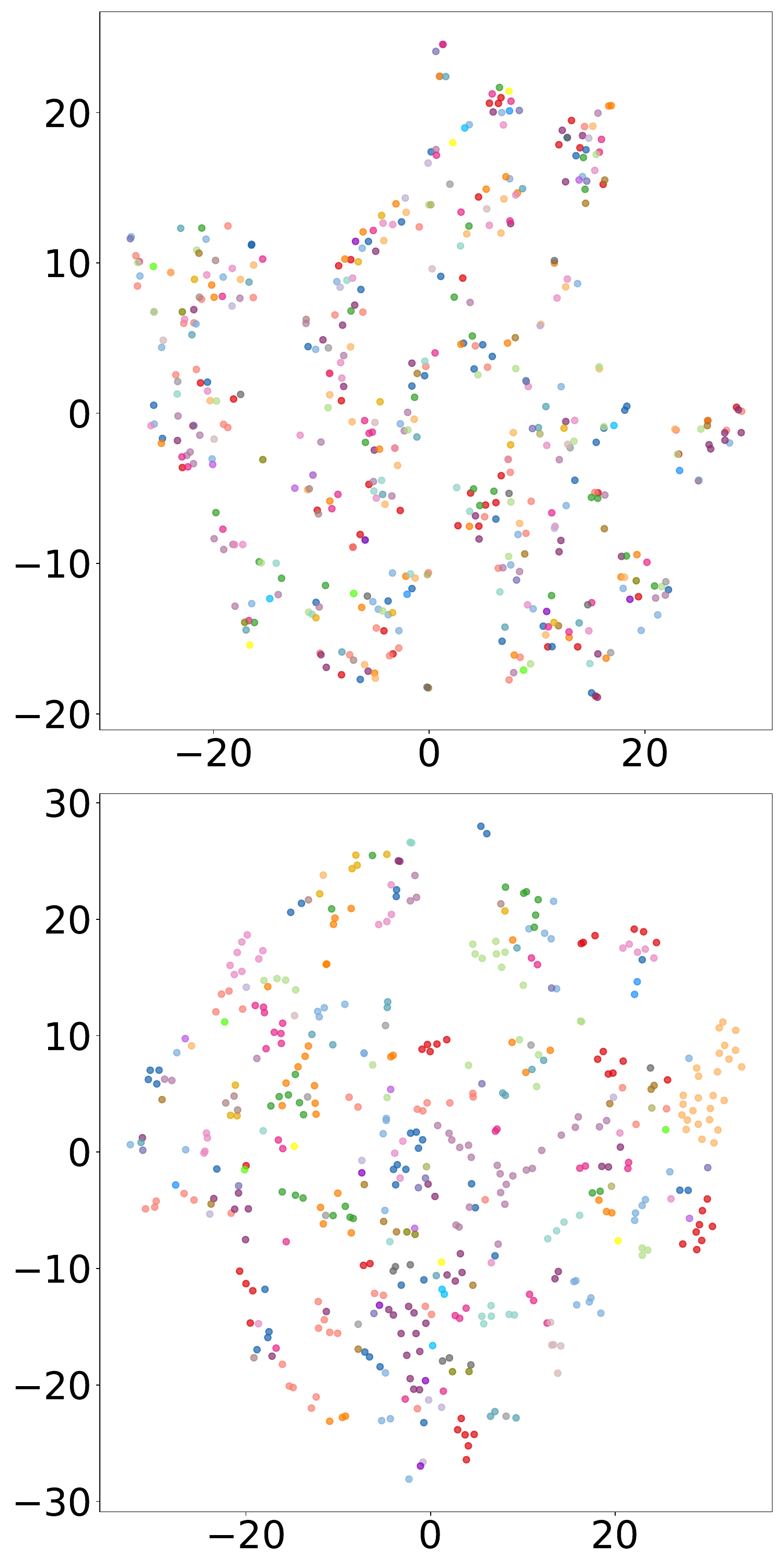}
        \end{minipage}%
    }\hfill
    \subfloat[DSN-DDI]{%
        \begin{minipage}[b]{0.2\linewidth}
            \includegraphics[width=\linewidth]{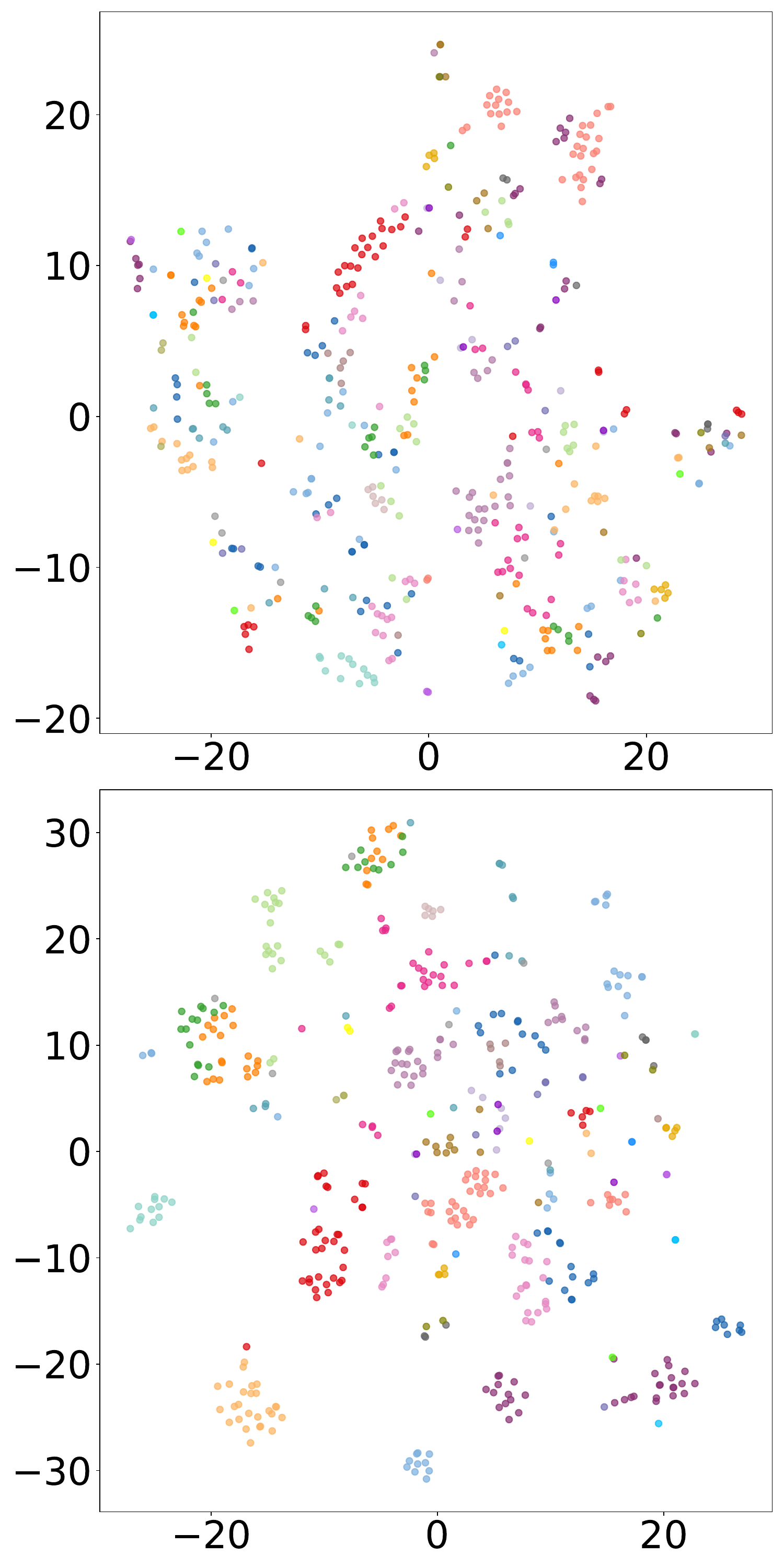}
        \end{minipage}%
    }\hfill
    \subfloat[MolBridge]{%
        \begin{minipage}[b]{0.2\linewidth}
            \includegraphics[width=\linewidth]{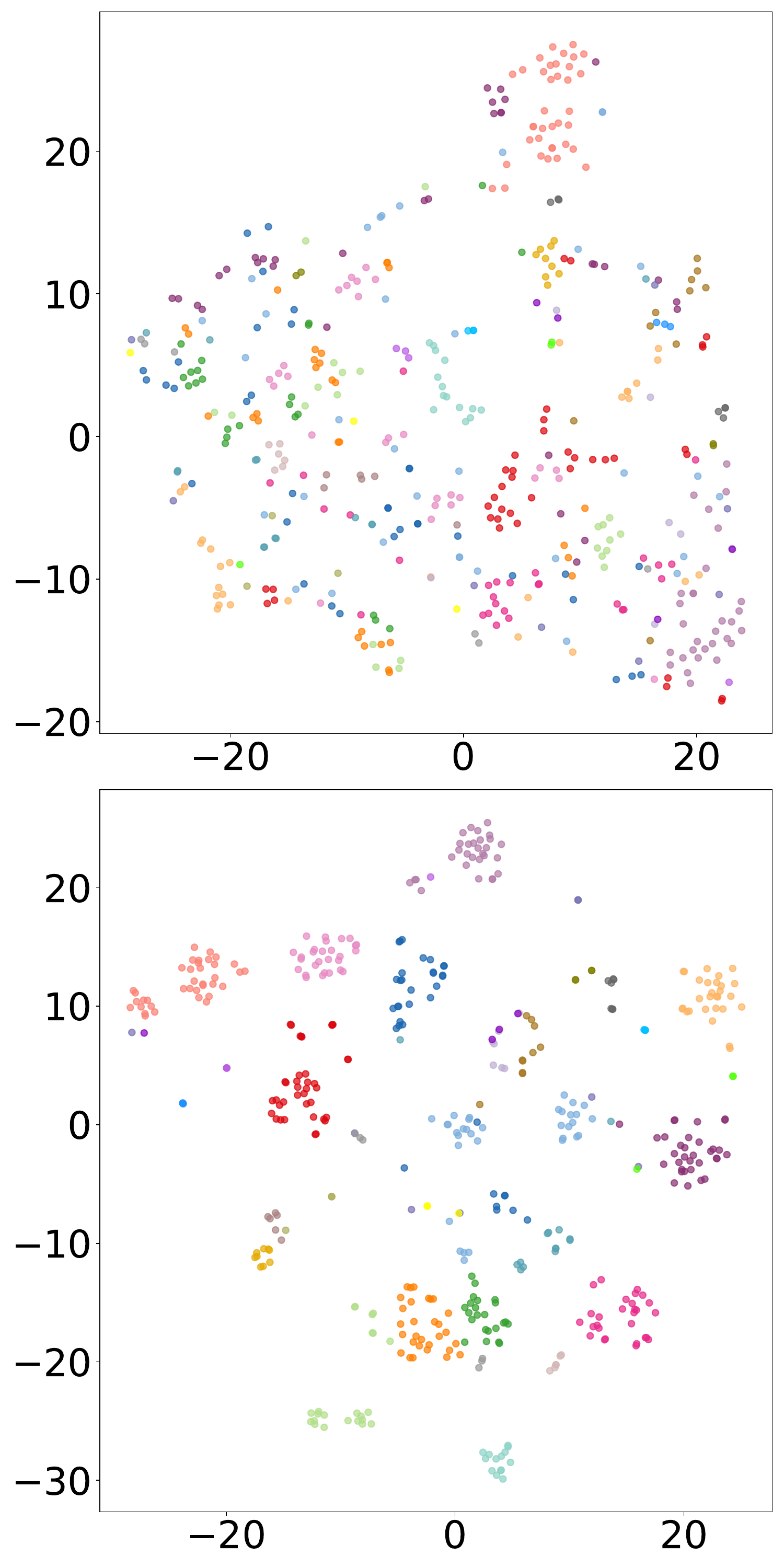}
        \end{minipage}%
    }
    \caption{The t-SNE visualization of 30 least-frequent events on Deng's dataset. Each point represents a pair of drugs, and the color denotes the type of DDI event. Upper: Initial feature encoding of drug pairs. Lower: Embedded representation of drug pairs.}
    \label{fig:d-tsne-han}
\end{figure*}

In addition to visualizing the embedded representations of drug pairs, we also visualized the feature encodings of drug pairs on Deng's dataset to facilitate vertical comparison. We reproduced models such as TrimNet, SSI-DDI, MRCGNN, and DSN-DDI, and used their original settings. 
\textcolor{black}{Figure \ref{fig:d-tsne-han} shows the visualization of the thirty DDI events with the lowest occurrence frequency in the samples. The visualization of the four most-frequent events is available in Supplementary Material Section 4 (Figure 5).} In each figure, the top row presents the visualization of the initial feature encoding of drug pairs obtained by concatenating the initial feature encodings of drugs for each model, and the bottom row shows the visualization of the embedded representation of drug pairs, with the same meaning as in the main text. Through vertical comparison, it can be found that MolBridge more effectively transforms the originally scattered feature representations of drug pairs into compactly distributed embedded representations of drug pairs. This indicates that MolBridge effectively extracts high-quality representations of drug pairs by integrating the local structural information of molecules and the long-distance implicit associations of atoms.
% Figure \ref{fig:d-tsne-duo} shows the visualization of the four DDI events with the highest occurrence frequency in the samples, and Figure \ref{fig:d-tsne-han} shows the visualization of the thirty DDI events with the lowest occurrence frequency in the samples. 

% \begin{figure*}[h]
%     \centering
%     \subfloat[TrimNet-DDI]{%
%         \begin{minipage}[b]{0.2\linewidth}
%             \includegraphics[width=\linewidth]{figs/tsne/t_sne_TrimNet_duo_.pdf}
%         \end{minipage}%
%     }\hfill
%     \subfloat[SSI-DDI]{%
%         \begin{minipage}[b]{0.2\linewidth}
%             \includegraphics[width=\linewidth]{figs/tsne/t_sne_SSI_duo_.pdf}
%         \end{minipage}%
%     }\hfill
%     \subfloat[MRCGNN]{%
%         \begin{minipage}[b]{0.2\linewidth}
%             \includegraphics[width=\linewidth]{figs/tsne/t_sne_MRCGNN_duo_.pdf}
%         \end{minipage}%
%     }\hfill
%     \subfloat[DSN-DDI]{%
%         \begin{minipage}[b]{0.2\linewidth}
%             \includegraphics[width=\linewidth]{figs/tsne/t_sne_DSN_duo_.pdf}
%         \end{minipage}%
%     }\hfill
%     \subfloat[MolBridge]{%
%         \begin{minipage}[b]{0.2\linewidth}
%             \includegraphics[width=\linewidth]{figs/tsne/t_sne_2d_duo_.pdf}
%         \end{minipage}%
%     }
%     \caption{Visualization of the 4 most-frequent events using t-SNE on Deng's dataset. Each point represents a pair of drugs, and the color denotes the type of DDI event. Upper: Initial feature encoding of drug pairs. Lower: Embedded representation of drug pairs.}
%     \label{fig:d-tsne-duo}
% \end{figure*}

\section{Supplementary Case Studies.}

\begin{figure}[h]
\centering
\includegraphics[width=\textwidth]{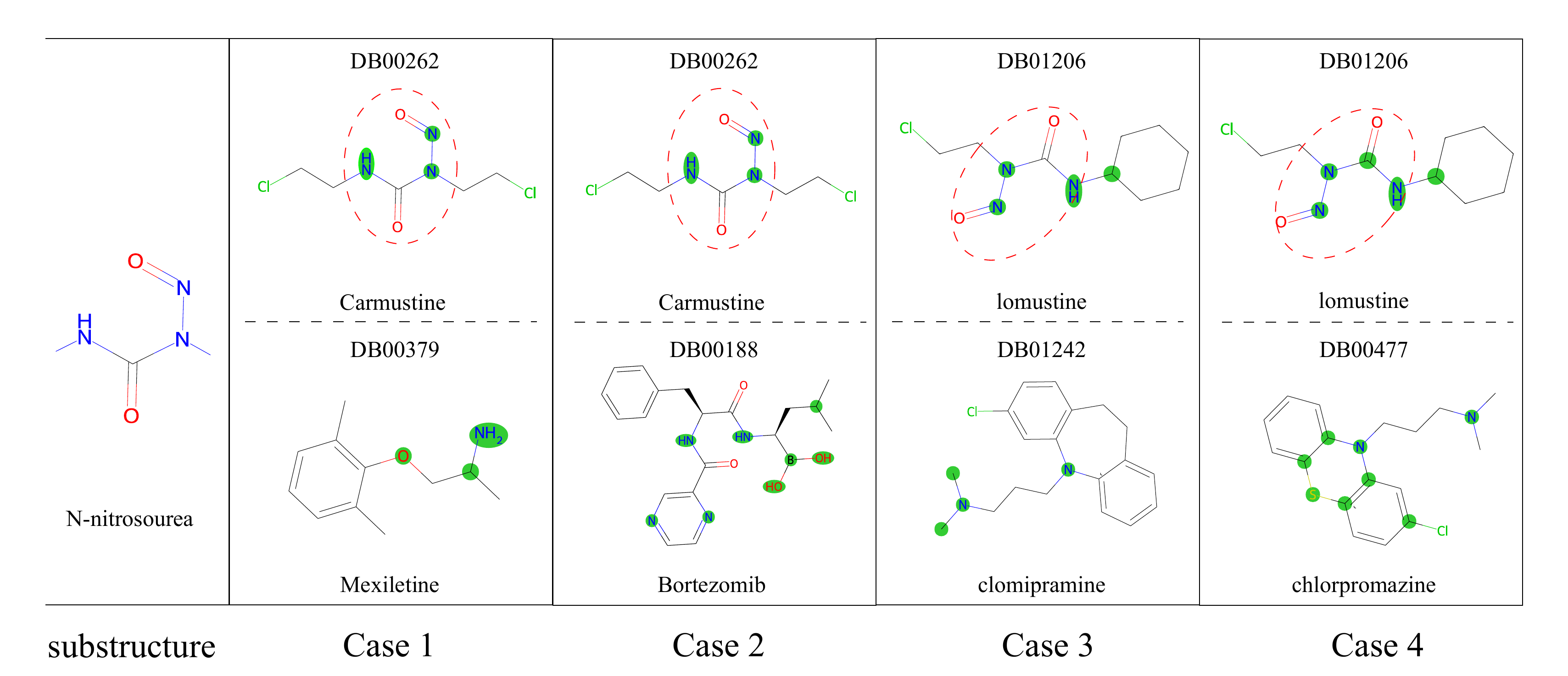}
\caption{Comprehensive drug interaction case studies showing MolBridge's identification of key molecular substructures. The figure displays four distinct interaction cases involving N-nitrosourea compounds (Carmustine and Lomustine) paired with drugs from different therapeutic classes. In each case, the highlighted substructures represent the model's prediction of critical interaction sites, consistently identifying N-nitrosourea groups as the primary interaction determinants across all drug pairs.}
\label{fig:comprehensive_cases}
\end{figure}

This appendix provides comprehensive experimental results demonstrating the robustness of our MolBridge framework across multiple drug interaction scenarios. Figure \ref{fig:comprehensive_cases} presents detailed analysis of four representative cases, showcasing the model's consistent ability to identify pharmacologically relevant interaction sites.
The extended case studies in Figure \ref{fig:comprehensive_cases} demonstrate that our framework maintains consistent performance across structurally and functionally diverse drug families. The universal identification of N-nitrosourea groups as interaction hotspots provides strong evidence for the mechanistic basis of these predictions, as these compounds are known alkylating agents that can interfere with drug metabolism through metabolic competition, covalent modification, and reactive intermediate formation. These findings validate MolBridge's utility for predicting clinically relevant drug-drug interactions in complex therapeutic regimens.

\section{Complete Inductive Evaluation Results}

Tables~\ref{tab:inductive_s1_app} and \ref{tab:inductive_s2_app} present comprehensive inductive evaluation results on DrugBank under S1 (one seen drug, one unseen) and S2 (both drugs unseen) settings. MolBridge achieves the best performance in both settings, surpassing CSMDD by +8.3/+0.7 Macro-F1 and +1.8/+3.0 Acc in S1/S2 respectively. Methods marked with $^*$ use external knowledge graphs.

\begin{table}[h]
    \centering
    \caption{S1 inductive evaluation on DrugBank (one seen drug, one unseen)}
    \label{tab:inductive_s1_app}
    \begin{tabular}{lccc}
        \toprule
        \textbf{Methods} & \textbf{Macro-F1} & \textbf{Acc} & \textbf{Kappa} \\
        \midrule
        MLP & 21.1±0.8 & 46.6±2.1 & 33.4±2.5 \\
        Similarity & 43.0±5.0 & 51.3±3.5 & 44.8±3.8 \\
        CSMDD & \underline{45.5±1.8} & \underline{62.6±2.8} & \underline{55.0±3.2} \\
        STNN-DDI$^*$ & 39.7±1.8 & 56.7±2.6 & 46.5±3.4 \\
        HIN-DDI$^*$ & 37.3±2.9 & 58.9±1.4 & 47.6±1.8 \\
        MSTE$^*$ & 7.0±0.7 & 51.4±1.8 & 37.4±2.2 \\
        KG-DDI$^*$ & 26.1±0.9 & 46.7±1.9 & 35.2±2.5 \\
        CompGCN$^*$ & 26.8±2.2 & 48.7±3.0 & 37.6±2.8 \\
        Decagon$^*$ & 24.3±4.5 & 47.4±4.9 & 35.8±5.9 \\
        KGNN$^*$ & 23.1±3.4 & 51.4±1.9 & 40.3±2.7 \\
        SumGNN$^*$ & 35.0±4.3 & 48.8±8.2 & 41.1±4.7 \\
        DeepLGF$^*$ & 39.7±2.3 & 60.7±2.4 & 51.0±2.6 \\
        \textbf{MolBridge} & \textbf{53.8±5.0} & \textbf{64.4±2.8} & \textbf{56.9±3.7} \\
        \bottomrule
    \end{tabular}
\end{table}

\begin{table}[h]
    \centering
    \caption{S2 inductive evaluation on DrugBank (both drugs unseen)}
    \label{tab:inductive_s2_app}
    \begin{tabular}{lccc}
        \toprule
        \textbf{Methods} & \textbf{Macro-F1} & \textbf{Acc} & \textbf{Kappa} \\
        \midrule
        CSMDD & \underline{19.8±3.1} & \underline{37.3±4.8} & \underline{22.0±4.9} \\
        HIN-DDI$^*$ & 8.8±1.0 & 27.6±2.4 & 13.8±2.4 \\
        KG-DDI$^*$ & 1.1±0.1 & 32.2±3.6 & -- \\
        DeepLGF$^*$ & 4.8±1.9 & 31.9±3.7 & 8.2±2.3 \\
        \textbf{MolBridge} & \textbf{20.5±1.3} & \textbf{40.3±2.7} & \textbf{25.9±3.0} \\
        \bottomrule
    \end{tabular}
\end{table}

\section{Space Complexity Analysis}
  Let $d$ denote the input atomic feature dimension before projection, $dim$ the projected hidden dimension, $d_{\mathrm{hid}}$ the FFN hidden size, $L$ the number of GFormer layers, and $C$ the number of DDI event classes. The parameter count mainly comes from: (i) feature projection $\mathcal{O}(d \cdot dim)$, (ii) multi-head attention over joint node features $\mathcal{O}(dim^2)$ (query/key/value projections and the output projection), (iii) $L$ GFormer layers with FFN parameters $\mathcal{O}(L \cdot dim \cdot d_{\mathrm{hid}})$, and (iv) the output classifier $\mathcal{O}(dim \cdot C)$. The dominant term is typically $\mathcal{O}(L \cdot dim \cdot d_{\mathrm{hid}})$ when $d_{\mathrm{hid}}$ is sufficiently large.

\end{document}